\documentclass[lettersize,journal]{IEEEtran}

\usepackage{xcolor}  

\usepackage{amssymb}

\usepackage{amsmath}
\usepackage{mathdots}

\usepackage{mathtools}

\usepackage{tensor}

\usepackage{urwchancal}
\DeclareFontFamily{OT1}{pzc}{}
\DeclareFontShape{OT1}{pzc}{m}{it}{<-> s * [1.10] pzcmi7t}{}
\DeclareMathAlphabet{\mathpzc}{OT1}{pzc}{m}{it}

\usepackage{graphicx}
\usepackage{ifpdf}
\ifpdf
\usepackage{epstopdf}
\PrependGraphicsExtensions{.svg}
\fi

\providecommand{\R}{\mathbb{R}}
\providecommand{\GP}{\mathbf{N}} 

\DeclareMathOperator{\diag}{diag}

\providecommand{\td}{\mathrm{d}}

\usepackage{accents}
\usepackage{mathtools}
\makeatletter
\providecommand{\scirc}{%
    \hbox{\fontfamily{\rmdefault}\fontsize{0.4\dimexpr(\f@size pt)}{0}\selectfont{\raisebox{-0.52ex}[0ex][-0.52ex]{$\circ$}}}}

\makeatother

\makeatletter
\providecommand{\ucirc}{%
    \hbox{\fontfamily{\rmdefault}\fontsize{0.4\dimexpr(\f@size pt)}{0}\selectfont{\raisebox{0.0ex}[0ex][-0.52ex]{$\circ$}}}}

\makeatother

\mathchardef\mhyphen="2D

\providecommand{\etal}{\textit{et al.}~}

\usepackage{amsmath}
\usepackage{amsfonts}
\usepackage{algorithmic}
\usepackage{algorithm}
\usepackage{array}
\usepackage[caption=false,font=normalsize,labelfont=sf,textfont=sf]{subfig}
\usepackage{textcomp}
\usepackage{stfloats}
\usepackage{url}
\usepackage{verbatim}
\usepackage{graphicx}
\usepackage{bbding}
\usepackage{hyperref}

\usepackage{wrapfig}

\providecommand{\etal}{\textit{et al}.}

\providecommand{\eg}{\textit{e}.\textit{g}., }

\usepackage{array}
\usepackage{booktabs}
\usepackage{mathrsfs}

\usepackage{xcolor}
\DeclareMathAlphabet{\mathcal}{OMS}{cmsy}{m}{n}
\SetMathAlphabet{\mathcal}{bold}{OMS}{cmsy}{b}{n}
\providecommand{\diag}{\textrm{diag}}
\providecommand{\diag}{\textrm{Cov}}

\begin{document}

\title{
	Asynchronous Blob Tracker for Event Cameras
}

\author{Ziwei Wang, Timothy Molloy, Pieter van Goor and Robert Mahony
\thanks{The authors are with the Systems Theory and Robotics (STR) Group, College of Engineering and Computer Science, Australian National University, Canberra, ACT 2601, Australia. Email: \{ziwei.wang1, timothy.molloy, pieter.vangoor, robert.mahony\}@anu.edu.au}
\thanks{Manuscript was accepted on August 7, 2024 by IEEE Transactions on Robotics. DOI: 10.1109/TRO.2024.3454410.}
\thanks{© 2024 IEEE.  Personal use of this material is permitted.  Permission from IEEE must be obtained for all other uses, in any current or future media, including reprinting/republishing this material for advertising or promotional purposes, creating new collective works, for resale or redistribution to servers or lists, or reuse of any copyrighted component of this work in other works.}
}



\maketitle

\begin{abstract}
Event-based cameras are popular for tracking fast-moving objects due to their high temporal resolution, low latency, and high dynamic range.
In this paper, we propose a novel algorithm for tracking event blobs using raw events \emph{asynchronously} in real time.
We introduce the concept of an \emph{event blob} as a spatio-temporal likelihood of event occurrence where the conditional spatial likelihood is blob-like.
Many real-world objects such as car headlights or any quickly moving foreground objects generate event blob data.
The proposed algorithm uses a nearest neighbour classifier with a dynamic threshold criteria for data association coupled with an extended Kalman filter to track the event blob state.
Our algorithm achieves highly accurate blob tracking, velocity estimation, and shape estimation even under challenging lighting conditions and high-speed motions ($>$ 11000 pixels/s).
The microsecond time resolution achieved means that the filter output can be used to derive secondary information such as time-to-contact or range estimation, that will enable applications to real-world problems such as collision avoidance in autonomous driving. 
Project Page: \href{https://github.com/ziweiwwang/AEB-Tracker}{{https://github.com/ziweiwwang/AEB-Tracker}}
\end{abstract}


\section{Introduction}
\begin{figure}[t!]
	\centering
	\begin{tabular}{c}
		\includegraphics[width=0.95\linewidth]{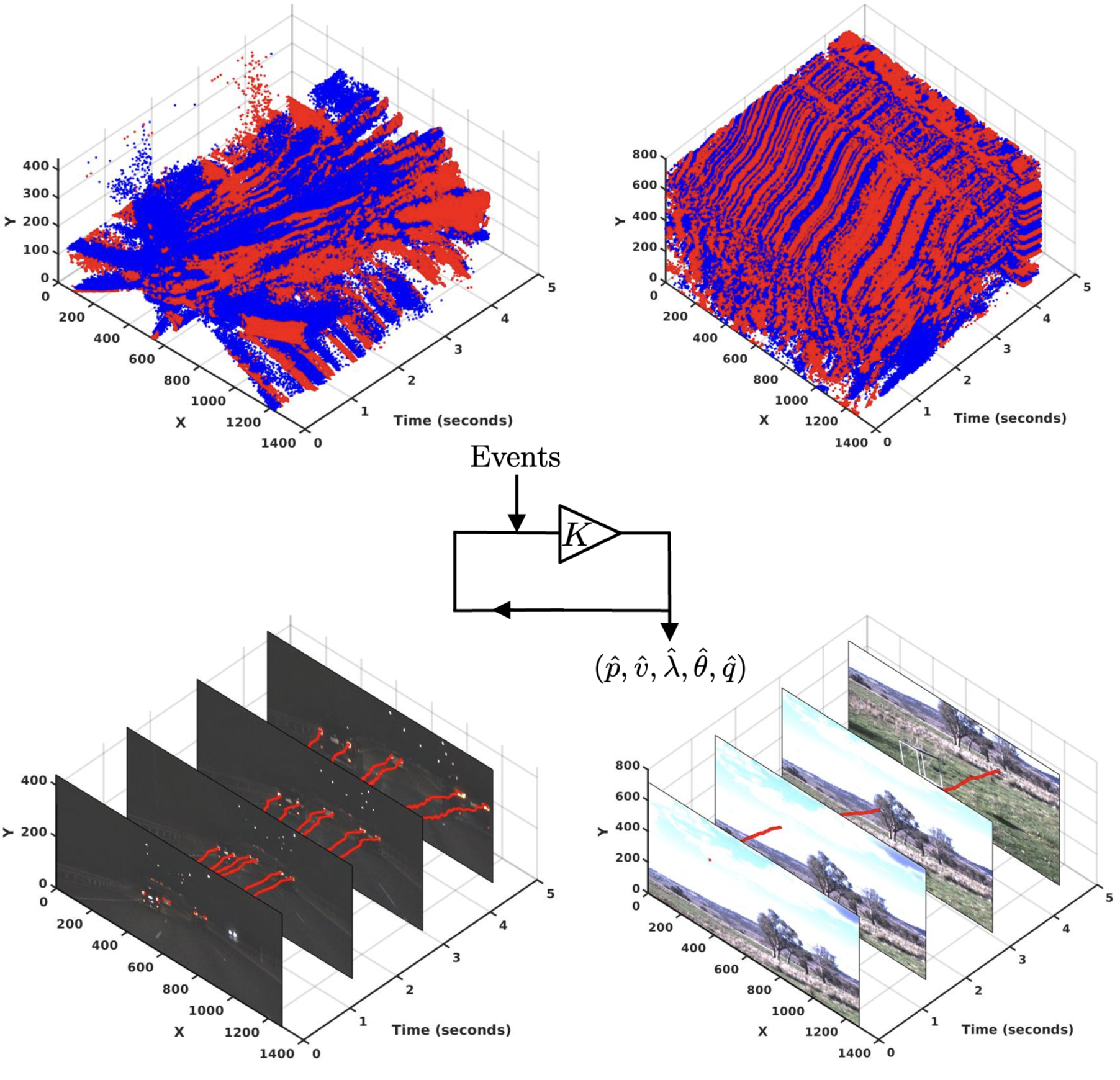}
	\end{tabular}
	\caption{\label{fig:front page}
Our asynchronous event blob tracker leverages asynchronous raw events (top row) to provide high-bandwidth estimates of an event blob's position ($p$), linear ($v$) and angular ($q$) velocity, orientation ($\theta$), and shape ($\lambda$).
The filter operates effectively in challenging scenarios including high-speed motions and extreme lighting conditions, and yields state updates up to microsecond resolution depending on event rates.
Potential applications include tracking multi-vehicle tail lights (left bottom) or aerial vehicles (right bottom).
The input event data are shown in the first row.
The high data rate trajectory estimated by our algorithm is marked by red lines and they traverse through the 40Hz reference images along the time axis.
Note that images are blind between frames and are only used for visualisation.
The high temporal resolution and shape estimation provided by the filter enable downstream data processing such as time-to-contact (TTC) and range estimation.
}\end{figure}

\IEEEPARstart{O}{bject} tracking is a core capability in a wide-range of robotics and computer vision applications such as simultaneous localisation and mapping (SLAM), visual odometry (VO), obstacle avoidance, collision avoidance, autonomous driving, virtual reality, smart cities, \textit{etc}~\cite{gallego2020event}.
Real-time visual tracking of high-speed targets in complex environments and in low light poses significant challenges due to their fast movement and complex visual backgrounds.
Traditional frame-based tracking methods are often hardware-limited as images are susceptible to motion blur, low frame rates and low dynamic range.
Event cameras offer significant advantages in these scenarios.
Instead of accumulating brightness within a fixed frame rate for all pixels, event cameras capture only changing brightness at a pixel-by-pixel level. Quickly moving foreground objects generate more events than complex visual backgrounds making the event camera sensor modality ideal for high-performance real-world target tracking.

In this paper, we propose an asynchronous event blob tracker (AEB tracker) for event cameras.
The targets considered are modelled as a spatio-temporal event likelihood where the conditional spatial likelihood is blob-like, a structure we term an \textbf{event blob} \cite{Martel18iscas}.
A key advantage is that the event blob concept fundamentally includes the temporal aspect of event data.
In particular, we use a spatial Gaussian likelihood with time-varying mean and covariance to model the unipolar event blob (where both polarities are equally considered).
The state considered includes position (mean of the event blob), non-homogeneous spatial cross-correlation of the event blob (encoded as two principal correlations and an orientation), and both the linear and angular velocity of the blob.
In the case of a non-flickering blob an additional polarity offset shape parameter is used as well.
Our AEB tracker uses an extended Kalman filter that accepts raw event data and uses each event directly to update a stochastic state-estimate asynchronously.
A key novelty of the algorithm is that the filter states for the shape of the distribution are used as stochastic parameters in the generative measurement model.
This formulation is non-standard in classical extended Kalman filters.
To do this we construct two pseudo measurement functions based on normalised position error and chi-squared variance.
These pseudo measurements provide information about the spatial variance of the event blob and ensure observability of the shape parameters of the target.
The proposed model enables adaptive estimation of blob shape allowing our AEB tracker to function effectively across a wide range of motion and shape tracking problems.
In particular, adaptive shape estimation allows our algorithm to operate without dependence on scenario-specific prior knowledge.

The asynchronous filter state is updated after each event (associated with a given blob), coupling the filter update rate to the raw event rate of the blob tracked.
For scenarios with low background noise, the foreground event rate is limited only by the microsecond physical limits of the camera
(data bus, circuit noise and refractory period limitations of the camera \textit{etc}~\cite{Wang2019}) and, even in poor lighting conditions, filter updates in excess of 50kHz are possible.

We evaluate the proposed algorithm on two indoor experiments, with extremely fast target speeds and camera motions; and two outdoor real-world case studies, including tracking automotive tail lights at night and tracking a quickly moving aerial vehicle.
In the first indoor experiment, we track a spinning blob with controllable speeds, ranging from low to very high optical flow (greater than 11000 pixels/s), and compare the proposed algorithm to the most relevant state-of-the-art event-based tracking algorithms.
In the second indoor experiment, we shake the camera in an unstructured pattern to demonstrate the tracking performance under rapid changes of direction of the image blobs.
In our first outdoor real-world case study, we track multiple automotive tail lights in a night driving scenario.
Due to the high event rates induced by the flickering LED lights, the filter achieves update rates of over 100kHz with high signal-to-noise ratio.
The quality of the filter data is validated by computing robust, high-bandwidth, time-to-contact estimates from the visual divergence between tail lights on a target vehicle.
In the second outdoor real-world case study, we track and estimate the event distribution of a high-speed aerobatic quadrotor in difficult visual conditions and in front of complex visual backgrounds.
The visual shape estimation of the target event blob is used to infer range information demonstrating the utility of the shape parameter estimation and quality of the filter estimate.

The primary contributions of the paper are:
\begin{itemize}
\item
A novel asynchronous event blob tracker for real-time high-frequency event blob tracking.

\item
A novel modification of extended Kalman filter theory to estimate shape parameters of event blob targets.

\item
Demonstration of the performance of the proposed algorithm on two experimental and two real-world case studies with challenging datasets.
\end{itemize}
The datasets and source code are provided open-source for future comparisons.

\section{Related Works}
The first high-speed, low-latency tracking algorithm for dynamic vision sensors was developed in 2006~\cite{Litzenberger06dspws}.
Its application in the robotic goalie task was studied in subsequent works~\cite{Delbruck07iscas, Delbruck13fns}.
Recently, event cameras have been used in a range of blob-like object tracking applications including star tracking~\cite{Chin19cvprw,Chin20wacv,ng2023asynchronous}; high-speed particle tracking~\cite{Ni12jm,Drazen11fluids,wang2020stereo}; and real-time eye tracking~\cite{angelopoulos2021event,ryan2021real}.
The tracking algorithms developed for these applications have, in the most part, been bespoke algorithms that exploit specific target, motion or background properties and have a limited ability to generalise to other scenarios.

General-purpose event-based tracking algorithms have focused on corner and template tracking.
The corner tracking algorithms can be categorised into window-based data association or clustering methods~\cite{Zhu17cvpr,Zhu17icra,Manderscheid19cvpr,hu2022ecdt}, asynchronous event-only methods~\cite{Clady15nn,Alzugaray18ral,barranco2018real,li2019fa,alzugaray2020haste,duo2021asynchronous}, and asynchronous methods that use hybrid event-frame data~\cite{Gehrig19ijcv,duo2021asynchronous}.
Template-based methods for event cameras tend to be highly dependent on the scenario.
Mueggler~\etal\cite{mueggler2015towards} used a stereo event camera to detect and track spherical objects for collision avoidance.
Mitrokhin~\etal\cite{Mitrokhin18iros} estimated relative camera motion from a spatio-temporal event surface, then segmented and tracked moving objects based on the mismatch of the local event surface.
Falanga~\etal\cite{falanga2019fast, falanga2020dynamic} demonstrated the importance of low latency for sense-and-avoid scenarios.
Their experiments used 10ms windowed event data and classical blob tracking algorithms.
Sanket~\etal\cite{sanket2020evdodgenet} developed an onboard neural network for the same sense-and-avoid scenario.
Rodriguez-Gomez~\etal\cite{rodriguez2020asynchronous} used asynchronous corner tracking and then clustered corners to create objects that are tracked.
Li~\etal\cite{li2019robust}  and Chen~\etal\cite{chen2019asynchronous} converted event streams into pseudo-frames and then tracked objects using frames.
Recent work has seen considerable effort in deep learning methods for object tracking~\cite{Ramesh18bmvc, chen2020end, sanket2020evdodgenet, zhang2022stnet, zhu2022learning, Messikommer23cvpr}.
A disadvantage of such methods is that they require a large amount of training data.
In addition, Convolutional Neural Network (CNN) algorithms rely on data windowing to create pseudo-frames, compromising low-latency and real-time performance \cite{chen2020end,Messikommer23cvpr}.
Other state-of-the-art networks, such as Spiking Neural Network (SNN) and Graph Neural Network (GNN) ~\cite{zhang2022stnet, zhu2022learning}, although promising in various aspects, are intractable to run in real-time at kHz rates due to their large scale and computational complexity.

The combination of conventional frames and event cameras has also been explored for object tracking~\cite{zhang2021object, iaboni2022event}, enabling the use of data association techniques derived from classical computer vision literature.
Active LED lights are commonly used to create event blob targets for experimental work.
M{\"u}ller~\etal\cite{muller2011miniature} used LED markers in a low-power embedded Dynamic Vision Sensor system and introduced two active LED marker tracking algorithms based on event counting and the time interval between events.
Censi~\etal\cite{Censi13iros} mounted LED markers on a flying robot and proposed a low-latency tracking method that associated LED markers based on the time interval between events and then tracked each LED light using a particle filter.
Wang~\etal\cite{wang22iros} used active LED lights for high-speed visual communication with conventional blob detection and tracking methods on pseudo-frames.
These LED tracking methods usually require the targets to blink in known or very high frequencies.
A characteristic of the schemes discussed is that data association is mostly built into an asynchronous pre-processing module, such as a corner detector, clustering algorithm or frame-based correlation; or the algorithm requires pseudo-frames, event-frames or estimation of an event surface over a window of data.
Such architectures lead to increased latency and reduced frequency response of target tracking, although many of the algorithms reviewed still achieve excellent results, especially when compared to classical frame-based image tracking.

\section{Problem Formulation}

In this section we formulate the problem of tracking event blob targets.

\subsection{Spatio-Temporal Gaussian Likelihood Model for Event Blob Targets}
Event cameras report the relative log intensity change of brightness for each pixel asynchronously.
We consider the likelihood $\ell(\xi,t, \rho|e_k)$ of an event $e_k$ occurring at pixel location $\xi \in \mathbb{R}^2$ at time $t \geq 0$ and with polarity $\rho \in \{\pm 1\}$.
We will term such a likelihood model an \emph{event blob} if the spatial distribution of the conditional likelihood $\ell(\xi, \rho | t, e_k)$ for a fixed time $t$ is blob-like.
We propose a spatio-temporal Gaussian event blob likelihood function
\begin{align}
\begin{split}
&\ell(\xi,t,\rho | e_k) \\
&:= \frac{\gamma(t)B(\rho | \xi, t)}{2 \pi \det(\Lambda(t))} \exp\left( - \frac{1}{2} (\xi - p(t))^\top \Lambda(t)^{-2} (\xi - p(t)) \right),
\end{split}
\label{eq:intensity_Lambda}
\end{align}
where $p(t) \in \mathbb{R}^2$ is the pixel location of the centre of the object and $\Lambda(t) \in \mathbb{R}^{2 \times 2}$ is the shape of the object (encoded as the principal square root of the second-order moment of the event spatio-temporal intensity).
The positive-definite matrix $\Lambda(t) > 0$ is written
\begin{align}
\Lambda(t) := R(\theta(t)) \begin{pmatrix} \lambda^1(t) & 0 \\ 0 & \lambda^2(t) \end{pmatrix} R^\top(\theta(t))
\label{eq:Lambda}
\end{align}
for principal correlations $\lambda^1(t), \lambda^2(t) > 0$ and orientation angle $\theta(t)$ where $R(\theta(t))$ is the associated rotation matrix.
The scalar $\gamma(t)$ encodes the temporal dependence of the likelihood associated with changing event rate and $B(\rho | \xi, t)$ denotes a binomial distribution associated with whether the event has polarity $\{\pm 1\}$ depending on position and time.
If $\gamma$ is integrable on a time interval of interest, then the spatio-temporal Gaussian event blob likelihood could be normalised to produce a probability density on this time interval.
In this paper, however, we will use the conditional likelihood $\ell(\xi,\rho | t, e_k)$, a spatial Gaussian distribution with time-varying parameters.
In the sequel, we will often omit the time index from the state variables $p = p(t)$, $\theta = \theta(t)$, \textit{etc}., to make the notation more concise.

\begin{figure}[t!]
	\centering
		\begin{tabular}{c}
			\includegraphics[width=1\linewidth]{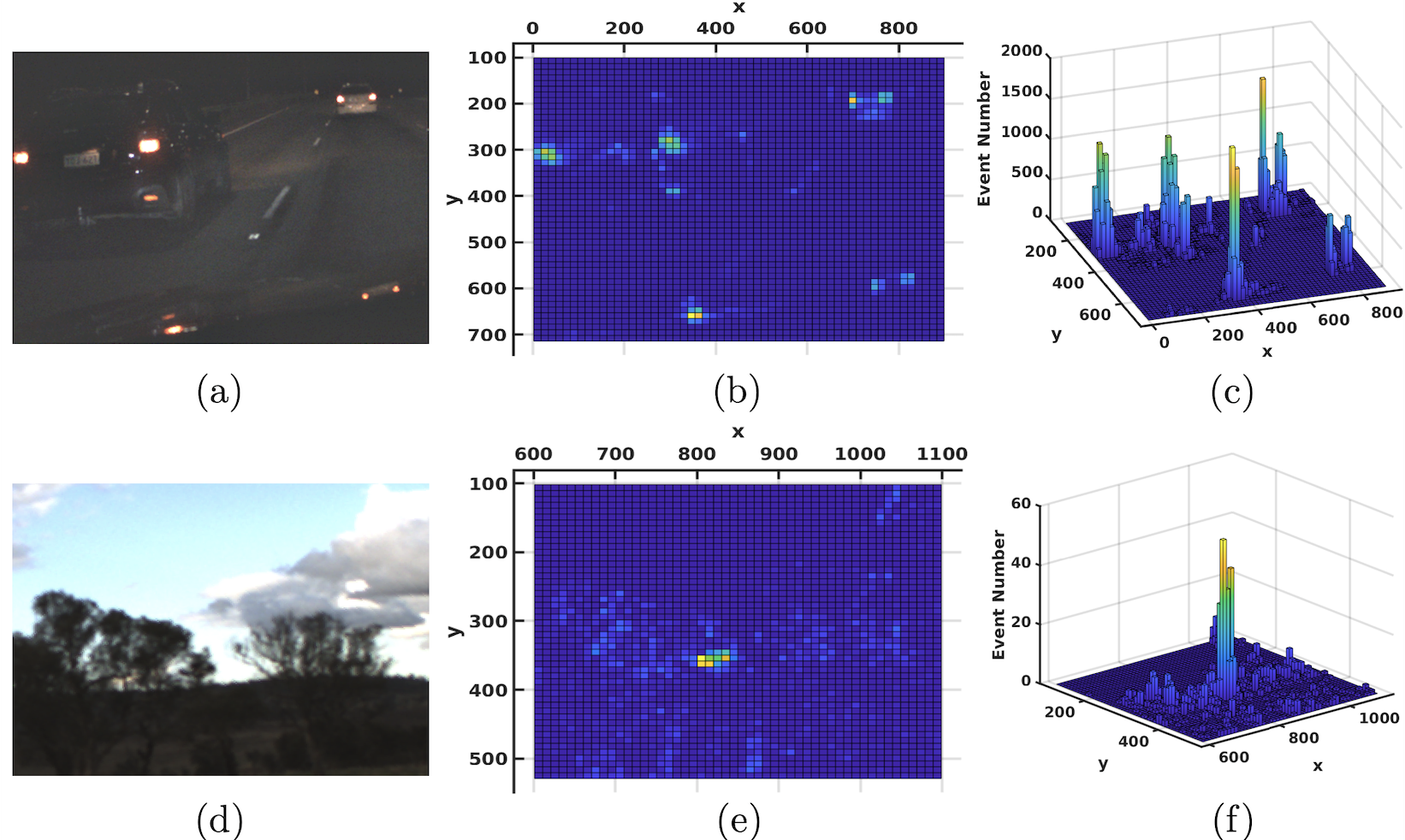}
		\end{tabular}
	\caption{\label{fig:Event-Histogram}
An example of plotting a short temporal window of events in a histogram pseudo-frame for the two case studies considered in the paper.
In the upper row, the LED car tail lights create a collection of event blobs.
(Note the reflections of the lights in the bonnet of the experimental car.)
In the bottom row, a quadrotor is flying in front of the trees and is difficult to pick out of the RGB image but generates an event blob that is clearly visible in (e) and (f).
The leading and trailing edges of the event blob can be seen in (e), while the overall blob structure is visible in (f).
}
\end{figure}

The archetypal examples of event blob targets are flickering objects such as an LED or fluorescent light~\cite{wang22icra}.
A flickering target generates events at pixel locations in proportion to the frequency of the flicker and intensity of the source at that pixel.
The likelihood of an event occurring at a given moment in time is proportional to the rate of events and the binomial probability of polarity is independent of position.
Thus, for a target with constant flicker, the spatial distribution of the unipolar (ignoring event polarity) event blob likelihood is directly related to the intensity of the source and is analogous to image intensity.
An example of the event arrival histogram for LED tail lights of cars is shown in the top row of Figure~\ref{fig:Event-Histogram}.
Here the image blobs in  Figure~\ref{fig:Event-Histogram}b-c are roughly Gaussian.

In the case of a non-flickering target, the situation is more complex since the motion of the target in the image plane is required to generate events.
In this case, the events are asymmetrically arranged around the target centre depending on whether the target is bright or dark with respect to the background and the binomial distribution $B(\rho|\xi, t)$ is not independent of the spatial parameter $\xi$.
For a bright target against a dark background, the events in the direction of motion are positive while the trailing events are negative, and vice-versa if the background is brighter than the target.
If both positive and negative events are equally considered, then the resulting density is symmetric around the centre point of the target (for an idealised sensor).
The rate of events at the target centre should be zero since the intensity of the target blob at this point is a maximum (or minimum) with respect to the direction of motion of the target.
The second row of Figure~\ref{fig:Event-Histogram} provides a good example of a non-flickering moving target.
The density of events in Figure~\ref{fig:Event-Histogram}e clearly shows the leading and trailing edge effects where the target is moving to the left and down in the image.
The effective likelihood over the whole target, however, appears to be a single blob as seen in Figure~\ref{fig:Event-Histogram}f.

In practice, we have found that a simple event blob model and the associated spatio-temporal Gaussian likelihood model proposed in \eqref{eq:intensity_Lambda} works well for a wide range of blob-like intensity targets.
For non-flickering targets (\eg Figure~\ref{fig:nonflicker}), we introduce a minor modification to the proposed algorithm (cf.~\S\ref{sec: Non-flickering Object Tracking}) to compensate for the bimodal distribution discussed above.

\subsection{Event Generation Model}
Consider a sequence of events ${e_k:=(\xi_k, t_k,\rho_k)}$ for k $\geq$ 0, at pixel locations $\{\xi_k\}$, times $\{t_k\}$ and with polarities $\{\rho_k\}$  associated with a target event blob modelled by \eqref{eq:intensity_Lambda}.
We will consider the unipolar model for event blobs where the polarity $\rho_k$ is ignored and all events are considered equally.
The associated likelihood is the marginal
\begin{align}
\ell(\xi,t|e_k) = \sum_{\rho_k \in \{\pm 1\}} \ell(\xi, t, \rho_k | e_k).
\end{align}
In addition, we will ignore noise in the time-stamp and consider the conditional likelihood $\ell(\xi| t = t_k, e_k)$  fixing time at $t = t_k$.
With modern event cameras, where the time resolution is as low as 1$\mu$s resolution this assumption is justified in a wide range of scenarios.
Based on these assumptions then a generative noise model for the event location is given by
\begin{align}\label{eq:event_gen_model}
\xi_k = p(t_k) + \Lambda(t_k) \eta_k, \quad \eta_k \sim \mathcal{N}(0, I_2),
\end{align}
where coordinate $p(t_k)$ is the true location of the object and we emphasise that the times $\{t_1, t_2, \ldots, t_k,\ldots\}$ are not periodic and depend on the asynchronous time-stamp of each event.
The measurement noise $\eta_k$ is a zero-mean independent and identically distributed (i.i.d.) Gaussian process with known covariance $I_2$. The noise process is scaled through the square-root covariance $\Lambda(t_k)$ given by
	\eqref{eq:Lambda} with parameters $\lambda^1(t_k), \lambda^2(t_k) > 0$ and $\theta(t_k)$.
\begin{figure*}[t!]
	\centering
	\includegraphics[width=0.75\linewidth]{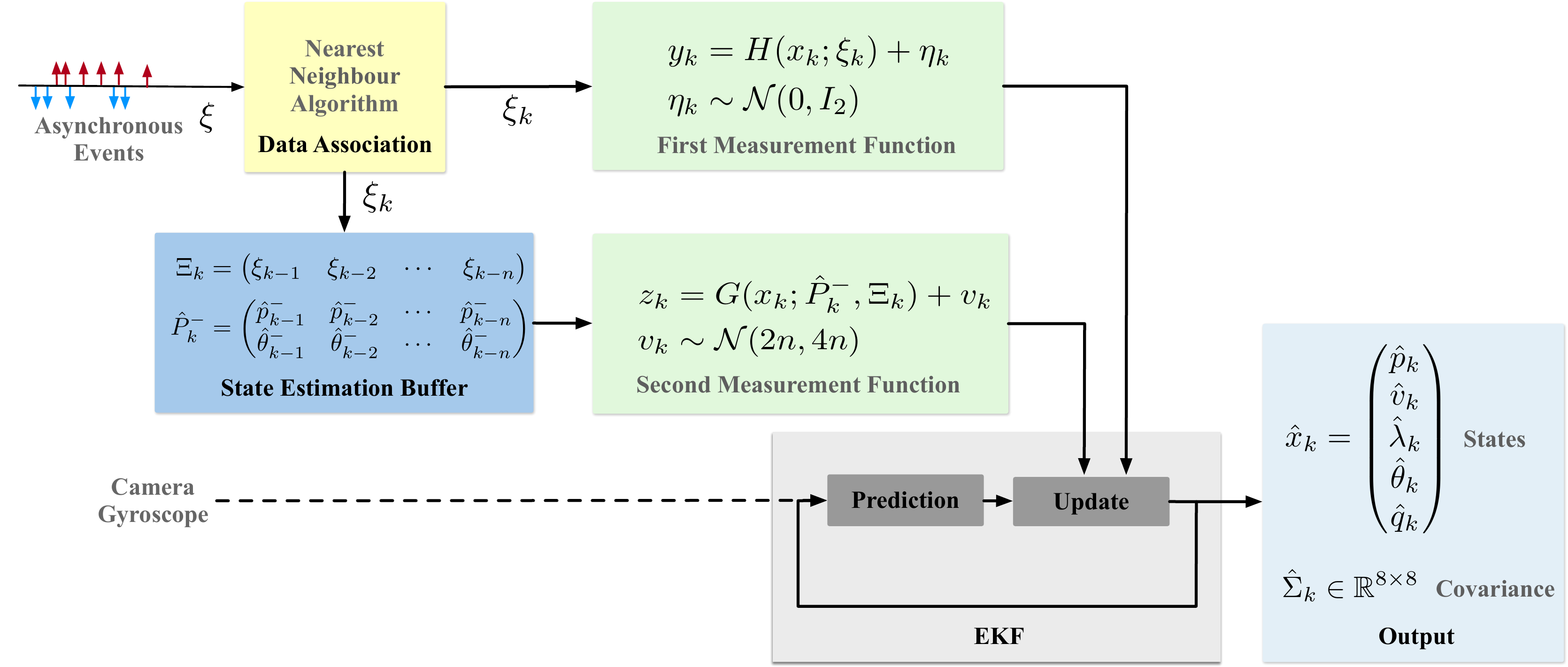} \\
	\caption{\label{fig:Pipeline}
		Block diagram of our asynchronous event blob tracking algorithm.
		The asynchronous filter state $\hat{x}_k$ updates with each event associated to a specific blob.
			Using the proposed two-stage measurement functions, the Extended Kalman filter (EKF) iteratively generates the optimal estimate for the state.
			The first function measures the normalised position error of the state, and the second function measures the variance of the chi-squared distribution formulated by the recent events and states saved in the state estimation buffer.
Predicting camera ego-motion using gyroscope angular velocity measurements is optional in the EKF and can be included when a calibrated IMU is available.
	}
\end{figure*}

\subsection{System Model}
For a 2D object in the image plane, the proposed model considers the states $p = (p_x, p_y)^\top$ for the object centre, and the shape parameters $(\lambda^1, \lambda^2)^\top$ and $\theta$ at pixel-level.
The parameter $\theta$ can be thought of as a spatial state parameter describing the orientation of the object, while the principal correlations
$\lambda = (\lambda^1, \lambda^2)^\top$ encode the eigenvalues of the second-order `visual' moment of the object.
We add a linear velocity state $v = (v_x, v_y)^\top$ and angular velocity $q$ for the spatial states $(p_x,p_y, \theta)$.
The state vector $x$ is
\begin{align}\label{eq:state_xi}
x := (p, v, \theta, q, \lambda) \in \R^8.
\end{align}
If an Inertial Measurement Unit (IMU) is available on the camera then the gyroscope measurement $\Omega = (\Omega_x, \Omega_y,\Omega_z)^\top$, can be used to provide feed-forward prediction for the state evolution in $p$ and $\theta$.
Define the matrix
	\begin{align}
    J & := \begin{pmatrix} 0 & \Omega_z \\  -\Omega_z & 0 \end{pmatrix} \label{eq:J}
\end{align}
and the partial state function~\cite[Chapter 15.2.1]{corke2011robotics}
\begin{align}
	f_p(p,\Omega) &:= \begin{pmatrix}
	\Omega_x \bar{p}_x \bar{p}_y / f  -\Omega_y (f + \bar{p}_x^2 / f) + \Omega_z \bar{p}_y \\
	\Omega_x(f + \bar{p}_y^2 / f)  -\Omega_y \bar{p}_x \bar{p}_y / f  - \Omega_z\bar{p}_x
	\end{pmatrix} \label{eq:f_p}
\end{align}
where $\bar{p}_x = (p_x - p_{x}^0$) and $\bar{p}_y = (p_y - p_{y}^0$) are the pixel coordinates relative to the principal point and $f$ is the camera focal length.
The proposed state model is
\begin{subequations}
\label{eq:ODE}
\begin{align}
\td p & = (v(t) + f_p(p,\Omega)) \td t  + (Q^p)^{\frac{1}{2}} \td w_p, \\
\td v & = J v(t) \td t + (Q^v)^{\frac{1}{2}} \td w_v ,  \\
\td \theta  & =(q(t) - \Omega_z) \td t +  (Q^\theta)^{\frac{1}{2}} \td w_\theta,  \\
\td q  & = (Q^q)^{\frac{1}{2}} \td w_q , \\
\td \lambda & =  (Q^\lambda)^{\frac{1}{2}} \td w_\lambda,
\end{align}
\end{subequations}
where we model uncertainty in the state evolution using continuous Wiener processes $w_p$, $w_v$, $w_\theta$, $w_q$ and $w_\lambda$ with $Q^p, Q^v, Q^\lambda \in \mathbb{R}^{2\times 2}$ and $Q^\theta, Q^q \in \mathbb{R}^{1\times 1}$ symmetric positive definite matrices.

Image blob velocity is modelled as a filter state, while change in velocity (acceleration) is compensated for in the process noise.
Camera angular velocity data $\Omega$ is the major source of ego-motion induced optical flow for image tracking tasks.
Ego-motion associated with camera IMU accelerometer measurements is not modelled in the present work as it is typically insignificant compared to the rotational ego-motion and independent motion of a target.
In addition, since there is often no reliable linear velocity sensor available on a real-world robotic system, the feature velocity estimation provided by the AEB tracker may be the best available sensor modality to estimate linear motion of the camera.
Due to its low latency and high temporal resolution, the feature velocity estimate of the AEB tracker is ideally suited to be used in this way. 
Even if the cameras linear velocity is available, it is necessary to know or estimate the range of the feature to include feature kinematics in the filter, introducing additional complexity.
The benefit of such an approach is unclear since the AEB tracker can easily track all but the most extreme motion without including feature kinematics.

\section{Asynchronous Event Blob Tracker}
\label{sec:ABtracker}
In this section, we present the proposed EKF-based asynchronous event blob tracker for event cameras (cf.~Figure~\ref{fig:Pipeline}).
We first present the EKF prediction step, before describing the measurement functions and the EKF update step.
We will discuss the data association and other implementation details in the next section.

Recall the continuous-time system state (Equation~\ref{eq:state_xi}) $x = (p, v, \theta, q, \lambda)$  for $t \in [0, \infty)$.
Define $(\hat{x}, \hat{\Sigma})$ to be the estimated state and covariance of $x \sim \GP(\hat{x}, \hat{\Sigma})$.
The prediction step of our tracker is computed as a continuous-time diffusion associated with the stochastic differential equation \eqref{eq:ODE}, while the update step of our tracker is an asynchronous update based on pseudo measurements constructed from event data and undertaken when each new event becomes available, similar to \cite{wang2021asynchronous, wang2023asynchronous}.

\subsection{Asynchronous Object Dynamics and Prediction Step}
The predicted state $\hat{x}$ is computed by integrating the dynamics \eqref{eq:ODE} without noise.
We linearise the dynamics \eqref{eq:ODE} about the predicted state $\hat{x}$ in order to predict the covariance $\hat{\Sigma}$.

Linearising $f_p$ in Equation~\eqref{eq:f_p} about $\hat{p}$ (the position components of $\hat{x}$) yields
\begin{align*}
A_{\hat{p}} = \begin{pmatrix}
\bar{\hat{p}}_y\Omega_x/f  - 2 \bar{\hat{p}}_x \Omega_y/f & \Omega_z + \bar{\hat{p}}_x \Omega_x /f \\
-\Omega_z -\bar{\hat{p}}_y \Omega_x/f & 2 \bar{\hat{p}}_y \Omega_x / f - \bar{\hat{p}}_x \Omega_y/f
\end{pmatrix}.
\end{align*}
It follows that the dynamics of the true state $x$, linearised about the estimated state $\hat{x}$, are
\begin{align}\label{eq:dynamics}
	\td x &\approx A (x - \hat{x}) \td t + Q^{\frac{1}{2}}  \td w,
	\end{align}
	where $w$ is Wiener process with positive definite matrix
	\begin{align} Q = \diag
	\begin{pmatrix} Q^p & Q^v & Q^\theta & Q^q & Q^\lambda \end{pmatrix},
\end{align}
and the state matrix $A = A(t)$ is a block matrix corresponding to the state partition of \eqref{eq:state_xi},
\begin{align}
A =
\begin{pmatrix}
A_{\hat{p}} & I_2 & 0 & 0 & 0 \\
0 & J & 0 & 0 & 0 \\
0 & 0 & 0 & 1 & 0 \\
0 & 0 & 0 & 0 & 0 \\
0 & 0 & 0 & 0 &  0
\end{pmatrix} \in \R^{8 \times 8},
\end{align}
with $J$ as defined in \eqref{eq:J}.
Note that, if $\Omega$ is unavailable or zero (the camera is stationary), then $A_{\hat{p}} = 0_{2\times 2}$.
In this case, the dynamics \eqref{eq:dynamics} are exactly linear and are thus independent of the linearisation point $\hat{x}$; that is, $\td x = A (x - \hat{x}) \td t + \td w$ and $A$ does not depend on $\hat{x}$.

The state covariance $\hat{\Sigma}$, in the absence of measurements, is predicted as the solution of a continuous-time linear Gaussian diffusion process given by
\begin{align} \label{eq: state covariance pred}
	\dot{\Sigma} = A \Sigma + \Sigma A^\top + Q,
\end{align}
where $A$ is computed around $\hat{x}$.

Let $\{t_k\}$ be a sequence of times associated with an event stream.
For each given time $t_k$, define the asynchronous system-state to be
\begin{align}
		x_k = (p_k, v_k, \theta_k, q_k, \lambda_k) := x(t_k)
\label{eq:8-state_discrete}
\end{align}
Analogously, define asynchronous estimated state parameters $(\hat{x}_k, \hat{\Sigma}_k)$ at time $t_k$.
Given estimated state parameters at time $t_k$ the predicted state $\hat{x}_{k+1}^- := x(t_{k+1})$ at time $t_{k+1}$ is the solution of \eqref{eq:ODE} without noise for initial condition $x(t_k) = \hat{x}_k$.
The predicted covariance $\hat{\Sigma}_{k+1}^- := \Sigma (t_{k+1})$ at time $t_{k+1}$ is the solution of \eqref{eq: state covariance pred} for initial condition $\Sigma(t_k) = \hat{\Sigma}_k$.
The actual computation of the prediction step used for the tracker algorithm is detailed in \S\ref{sec:prediction_computation}.

\subsection{Two-Stage Pseudo Measurement Construction}
\label{sec:pseudo-measurements}
Let $\{e_k\}$ be a sequence of events.
The natural measurement available for event blob target tracking is the location $\xi_k$ of each event $e_k = (\xi_k,t_k,\rho_k)$, 
available at discrete asynchronous times $t_k$.
The associated generative noise model for the raw event location measurement is given by \eqref{eq:event_gen_model}.
However, this model cannot be used directly in the proposed event-blob target tracking filter since the square-root uncertainty $\Lambda_k$ in the generative noise model is itself a state to be estimated by the filter (i.e., determined by variables $\lambda_k, \theta_k$ in $x_k$).
Instead, we will derive a pseudo measurement with known measurement covariance and use the theory of Kalman filtering with constraints~\cite{julier2007kalman}.

Define a measurement function by
\begin{align}\label{eq:virtual1}
H(x_k; \xi_k) := \Lambda_k^{-1}(\xi_k-p_k),
\end{align}
where $H$ is a function of the state $x_k$ parametrised by the measurement $\xi_k$.
Recalling \eqref{eq:event_gen_model}, then by construction
\begin{align*}
H(x_k; \xi_k) & = \Lambda_k^{-1}(\xi_k-p_k)
= \eta_k,
\end{align*}
where recall $\eta_k \sim \mathcal{N}(0, I_2)$ is an independent and identically distributed (i.i.d.) Gaussian process with known covariance.
In particular, the expected value $E[\Lambda_k^{-1}(\xi_k-p_k)] = 0$.

Define a new measurement model
\begin{align}\label{eq:yk}
y_k & = H(x_k; \xi_k) + \eta_k, \quad \eta_k \sim \mathcal{N}(0, I_2)
\end{align}
with pseudo measurements $y_k \equiv 0 \in \mathbb{R}^2$.
The generative noise model \eqref{eq:yk} with measurements $y_k = 0$ has known stochastic parameters $\eta_k \sim \mathcal{N}(0,I_2)$ and can be used in a Kalman filter construction.

The measurement \eqref{eq:yk} is insufficient to provide observability of the full state $x_k$.
Intuitively, this can be seen by noting that a decreasing measurement error $\|y_k - \Lambda_k^{-1} (\xi_k - p_k) \|=\|\Lambda_k^{-1} (\xi_k - p_k) \|$ can be modelled either as indicating $p_k$ should be moved towards $\xi_k$ or that $\Lambda_k$ should be increased (see Figure~\ref{fig:ablation} in the ablation study Section~\ref{Ablation Study}).

To overcome this issue, we introduce an additional pseudo measurement specifically designed to observe the shape parameter $\Lambda_k$.
The approach is to construct a chi-squared statistic from a buffer of prior state estimates that allows estimation of the shape parameters $\lambda_k = (\lambda^1_k, \lambda^2_k)$.

Assume that the extended Kalman filter has been operating for $n$ asynchronous iterations, with $n$ events received for the target.
During initialisation we will run a bootstrap filter to generate an initial $n$ time-stamps ${t_k}$ to enable a warm start for the full filter.
For a fixed index $k$ and window length $n$, let $\hat{p}^-_{k-j}$ and $\hat{\theta}^-_{k-j}$ denote the state-estimate filter predictions based on measurements $\xi_{k-j}$, where $j = (1, \ldots, n)$.
Define $\hat{P}_k^-$ to be the buffer of state prediction estimates
\begin{align}\label{eq:buffer1}
\hat{P}_k^- := \begin{pmatrix}
\hat{p}_{k-1}^- & \hat{p}_{k-2}^- & \cdots & \hat{p}_{k-n}^- \\
\hat{\theta}_{k-1}^- & \hat{\theta}_{k-2}^- & \cdots & \hat{\theta}_{k-n}^-
\end{pmatrix},
\end{align}
and $\Xi_k$ to be the buffer of the associated events
\begin{align}\label{eq:buffer2}
\Xi_k := \begin{pmatrix}
\xi_{k-1} & \xi_{k-2} & \cdots & \xi_{k-n}
\end{pmatrix}.
\end{align}
For $j = (1, \ldots, n)$, define
\begin{align}
\hat{\Lambda}_{k-j} & :=  \mathcal{R}(\hat{\theta}_{k-j}^-) \diag[\lambda_k^1, \lambda_k^2] \mathcal{R}  (-\hat{\theta}_{k-j}^-),
\end{align}
where $(\lambda^1_k,\lambda^2_k)$ are the principal correlations at time $k$, while the prior filter states are used to rotate the image moment to the best estimate of its orientation.
Note that we do not use all the available data at time $k$ for the estimation of $\hat{\theta}_{k-j}^-$ since doing so would introduce undesirable stochastic dependencies.

Let $\beta \geq \| \Sigma^-_{\hat{p}_{k-j}}\|_2 $ be a constant estimate of an over-bound for the uncertainty in $\hat{p}_{k-j}$.
Define
\begin{align}\label{eq:chi}
\chi_{k-j} := \frac{1}{1 + \beta} \hat{\Lambda}_{k-j}^{-1}(\xi_{k-j}-\hat{p}_{k-j}^-).
\end{align}
Note that $E[\hat{p}_{k-j}^-] = p_{k-j}$ and $E[\hat{\Lambda}_{k-j}] = \Lambda_{k-j}$.
Replacing the estimates in \eqref{eq:chi} by their expected values yields a scaled version of the event generation model \eqref{eq:event_gen_model}.
Thus, the primary contribution to the uncertainty in $\chi$ will be a Gaussian distribution $\mathcal{N}(0,\frac{1}{1+\beta}I_2)$.
Variance in $\hat{p}_{k-j}^-$ is bounded by the estimated parameter $\beta > 0$.
Since this variance has a similar structure to the variance in $\xi_k$ it can be approximated as contributing additional uncertainty $\frac{\beta}{1+\beta} I_2$ to $\chi_{k-j}$.
The additional uncertainty in $\hat{\theta}_{k-j}^-$ is only present in the rotation matrix $\mathcal{R}(\hat{\theta}_{k-j}^-)$ and its contribution to uncertainty in $\chi$ is negligible.
Thus, the uncertainty in $\chi$ can be modelled by
\[
\chi_{k-j}
\sim \mathcal{N}\left( 0,\frac{1}{1+\beta}I_2 + \frac{\beta}{1+\beta}I_2 \right)
= \mathcal{N}(0, I_2).
\]

Define a new virtual output function $G$
\begin{align}\label{eq:G}
	 G(x_k; \hat{P}_k^-, \Xi_k):=
\sum_{j=1}^{n}\|\chi_{k-j}\|^2,
\end{align}
with dependence on $x_k$ through $\hat{\Lambda}_{k-j}^{-1}$ in $\chi_{k-j}$ (cf. Equation \eqref{eq:chi}).
Since $\chi_{k-j}$ are normally distributed with unit covariance, the output function $G(x_k; \hat{P}_k^-, \Xi_k)$ is the sum of the squares of $n$ independent 2-dimensional Gaussian random variables. Consequently, it follows a chi-squared distribution with an order of $2n$.
According to the central limit theorem~\cite{box1978statistics}, as $n$ becomes sufficiently large, the chi-squared distribution converges to a normal distribution with a mean of $2n$ and a variance of $4n$.
Based on this, we propose a pseudo measurement model
\begin{align}\label{eq:zk}
z_k  := G(x_k; \hat{P}_k^-, \Xi_k)  + v_k, \quad v_k &\sim \mathcal{N}(2n, 4n),
\end{align}
with pseudo measurements $z_k \equiv 2n$.

This development depends on the bound $\beta \geq \| \Sigma^-_{\hat{p}_{k-j}}\|_2 $.
In practice, the uncertainty in $\hat{p}_{k-j}$ is often much smaller compared to the uncertainty in the event position $\xi_k$ associated with the actual shape of the object.
Typically $\Lambda_k$ is 100 to 1000 larger than $\Sigma^-_{\hat{p}_k}$, corresponding to a $\beta \in [10^{-2}, 10^{-3}]$.
Although this bound makes little difference in practice, it provides theoretical certainty that the filter will not become overconfident.

Define the combined virtual measurement
\begin{align}
	m_k = \begin{pmatrix} y_k \\  z_k \end{pmatrix} \in \R^3.
	\label{eq:mk}
\end{align}
The expected value is $m_k = (0,0,2n)^\top$ for all $k$ by the definitions of $y_k$ and $z_k$.
Note that $y_k$ is a function of the (current) event $\xi_k$ and (current) state $x_k$, while $z_k$ is a function only of the (past) events and the (past) filter estimate before the update step in the Kalman filter.
It follows that the random variables in $y_k$ \eqref{eq:yk} and $z_k$ \eqref{eq:zk} are independent.
This leads to the generative noise model
\begin{align}
	m_k & = \begin{pmatrix}
		H(x_k; \xi_k) \\
		G(x_k; \hat{P}_k^-, \Xi_k)
	\end{pmatrix}
	+ \nu_k , & \nu_k \sim\mathcal{N}(0,R) \\
	R & =\begin{pmatrix} 1 & 0 & 0 \\ 0 & 1 & 0 \\ 0 & 0 & 4n \end{pmatrix}, \label{eq: R matrix}
\end{align}
where $H(x_k; \xi_k)$ and $G(x_k; \hat{P}_k^-, \Xi_k)$ are independent.

\subsection{EKF Update Step}

We linearise the two non-linear measurement models $y_k$ and $z_k$ \eqref{eq:yk}-\eqref{eq:zk} by computing the Jacobian matrices of the $H(x_k; \xi_k)$ and $G(x_k; \hat{P}_k^-, \Xi_k)$ measurement functions.
The linearised observation model $C_k$ for the state $x_k$ is written
\begin{align}
\begin{split}
C_k &= \begin{pmatrix}
C_k^H \\ C_k^G
\end{pmatrix} \in \mathbb{R}^{3 \times 8}.
\end{split}
\end{align}
The detailed partial derivatives are provided in Appendix~\ref{sec:app-Observation Model}.

From here, we follow the standard extended Kalman filter algorithm to compute the pre-fit residual Kalman filter gain $K_k$ to be
\begin{align}
K_k = \hat{\Sigma}_k^- C_k (C_k \hat{\Sigma}_k^- C_k^\top + R)^{-1}.
\end{align}
Finally, the updated state estimate and covariance estimate are given by
\begin{align}
\hat{x}_{k} = \hat{x}^-_k+ K_k \tilde{y}_k, \\
\hat{\Sigma}_k = (I-K_k C_k) \hat{\Sigma}_k^-.
\end{align}


\section{Implementation of Asynchronous Event Blob Tracker}


In this section, we present the track initialisation, data association logic and other implementation details of the proposed AEB tracker.

\subsection{Track Initialisation}\label{sec:initialisation}
Following the precedent set by our primary comparison algorithms (ACE~\cite{Alzugaray18threedv} and HASTE~\cite{alzugaray2020haste}), we use manual selection (click on screen) or predefined locations for specific  experimental studies to initialise tracks in the suite of experimental studies documented in Section~\ref{sec:experiments}.
In practice, the AEB tracker would be initialised using detection algorithms methodologies tailored to specific applications, such as high speed particle tracking~\cite{wang2020stereo} or eye tracking~\cite{ryan2021real}, which require specialised detection methodologies.
There are a range of potential algorithms in the literature that could be used for this purpose \cite{Alzugaray18ral,Zhu17icra}.
Further discussion of automated track initialisation is beyond the scope of the present paper.

We initialise targets with zero initial linear and angular velocity, and the initial target size is set to a value at least two times larger than the maximum expected blob size to make the initial transient of the filter more robust.

In addition to the initial states, the proposed filter requires estimates of process noise covariance and an initial prior for the state covariance.
The tuning of these parameters follows standard practice for Kalman filter algorithms, allowing for data-based methods like Expectation Maximisation \cite{shumway2000time, holmes2013derivation}.
Alternatively, manual tuning of process noise and prior covariances can be performed to align with the specific application and anticipated motion.
The covariance prior for blob location and size is determined based on an estimation of the error in the blob detection methodology.
The covariance for linear and angular velocity is set in an order of magnitude larger than the position estimate to allow the filter to quickly converge for these states.
In cases of highly erratic motion, the value of $Q$ can be increased to model stochastic variation in the linear optical flow of the target.
For less erratic motions, $Q$ can be decreased to enhance the algorithm's resilience to outliers and noisy data.
The better the predictive model of the motion of the target is, the smaller the value of $Q$ can be made.
In particular, when an IMU is present and ego-motion of the target for camera rotation can be estimated, this will tend to allow for significantly smaller $Q$ and improve outlier rejection and robustness of the tracking performance.
This is important since the ego-rotation of the camera generates the most spurious outlier events from the background texture.
Default parameters for the covariance prior and $Q$ that we have found to work well across a wide range of scenarios is documented in the companion code.

\subsection{Data Association}\label{sec:data-assoc}
Data association in visual target tracking is arguably one of the most difficult problems.
In this paper, we take a simple approach that considers all events in a neighbourhood of the event blob to be inliers.
As long as the signal-to-noise ratio of the number of events generated by an event blob target to the background noise events is high enough, the disturbance from background noise events will be small and the filter will maintain track.
Furthermore, any background noise events that are spatially homogeneous,
such as an overall change of illumination, will be distributed in the developed pseudo measurements and will have only a marginal effect on the filter tracking performance.
It is only when two objects generating a large number of events cross in the image that the algorithm may lose track.
This may be due to two targets crossing in the image, such as when the tail lights from one car occlude the tail lights of another car.
Or when the ego-motion of the camera causes a high-contrast background feature to cross behind a low-intensity target.
Measuring camera ego-motion by an IMU and predicting the target's position  in the corresponding direction helps to address the second case.

There remains the question of choosing the neighbourhood in which to associate inliers and outliers.
The size of this neighbourhood must be adapted dynamically to adjust for the changing size of the target in the image.
Our data association approach uses the nearest neighbour classifier with a dynamic threshold $\sigma$.
That is, any event $\xi_k$ that lies less than $\sigma$ pixels from the predicted state estimate $\hat{p}^-_k$ is associated with the target and used in the filter.
The threshold $\sigma$ is chosen as a low-pass version of $\max(\hat{\lambda}^1_k, \hat{\lambda}^2_k)$.
In continuous-time, this low-pass filer is written
\begin{align}\label{eq:ODE data associ}
\dot{\sigma}(t) = - \alpha \sigma(t) + b \alpha \max(\hat{\lambda}^1_k, \hat{\lambda}^2_k),
\end{align}
where $\alpha$ is the filter gain and $b$ is the desired ratio between the distance threshold and the estimated target size.
Integrating \eqref{eq:ODE data associ} over the time interval $(t_{k} - t_{k-1})$ for $\max(\hat{\lambda}^1_k, \hat{\lambda}^2_k)$ constant yields
\begin{align}
\begin{split}
\sigma_k &= \beta_k \sigma_{k-1} + b (1-\beta_k) \max(\hat{\lambda}^1_k, \hat{\lambda}^2_k), \\
\beta_k &:= \exp(-\alpha (t_{k} - t_{k-1})).
\end{split}
\end{align}
The gain $\alpha$ is chosen based on the expected continuous-time dynamics of the target in the image.


\subsection{Multi-target Tracking}
Our AEB tracker tracks multiple targets independently.
Each target is updated as an individual state asynchronously as events arrive.
This approach allows us to maintain separate and accurate tracking for multiple blob-like targets.
When a target is detected a separate state model $x = (p, v, \theta, q, \lambda) \in \mathbb{R}^{8}$  and the corresponding matrices (\eg $Q$, $\Xi$, \textit{etc}.,) is initialised for each new target as discussed in Section~\ref{sec:initialisation}.
Events are processed asynchronously and assigned to separate targets using our data association method as discussed in Section \ref{sec:data-assoc}.
These events are used to update the state of the associated target through our proposed algorithm.
Any event not associated with a target is discarded.

\subsection{Non-flickering Object Tracking} \label{sec: Non-flickering Object Tracking}
	\begin{figure}[t]
		\centering
		\begin{tabular}{c}
			\includegraphics[width=1\linewidth]{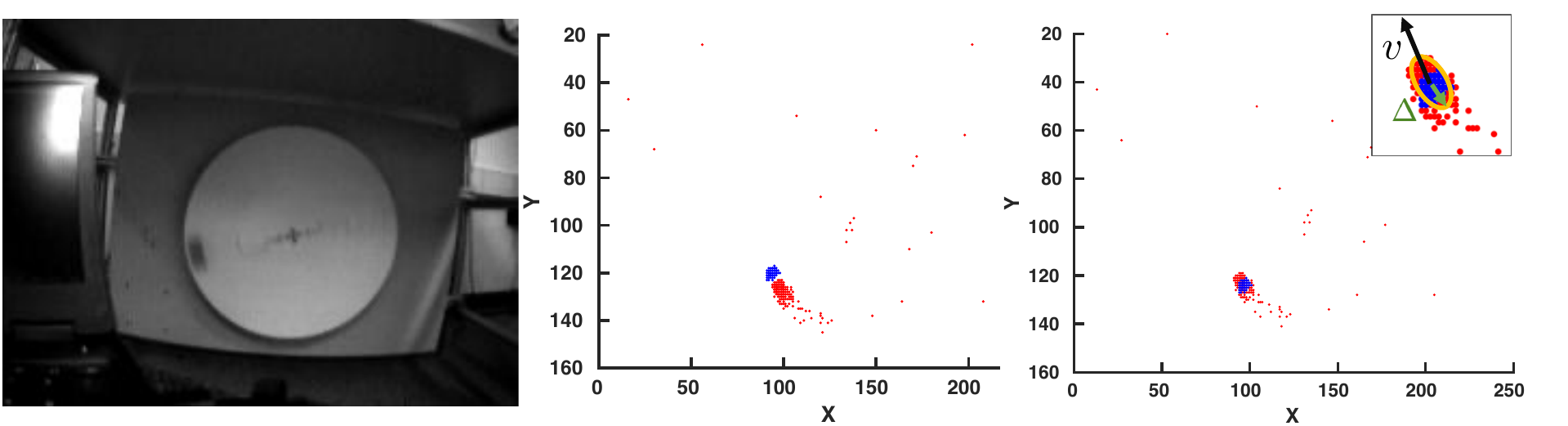}
		\end{tabular}
		\caption{\label{fig:nonflicker}
				Example of modelling the polarity offset vector $\Delta$ in the filter state to compensate the bimodal distribution of event data for
				tracking a non-flickering moving object.
				Left: an intensity frame showing a black target spinning clockwise.
				Middle: corresponding event data showing a bimodal distribution where the leading edge events are mostly negative (blue) and the trailing edge events are positive (red).
				Right:  event data distribution compensated by our polarity offset vector $\Delta$.
				The estimated velocity $v$ (black arrow), the polarity offset vector $\Delta$ (green arrow) and the estimated shape $\lambda$ (orange ellipse) are shown in the zoom in the right top corner of the right-most figure.
		}
	\end{figure}

Flickering targets such as LED lights or non-flickering drones with spinning rotors trigger events that can be approximated to a spatial-temporal Gaussian event blob (Figure \ref{fig:Event-Histogram}).
However, a moving non-flickering object triggers a bimodal distribution of event data where the leading edge events have one polarity while the trailing edge events have the opposite polarity, depending on the intensity contrast of the blob to the background (see Figure \ref{fig:nonflicker}).
In this case, an unmodified version of the AEB tracker algorithm tends to track either the leading or trailing edge of the target and may sometimes switch between the two edges causing inconsistent velocity estimation.
This behaviour only happens for non-flickering blob targets and arises because of the bimodal distribution of events.
	
To overcome this, we add a simple polarity offset parameter $\Delta = (\Delta_x, \Delta_y)^\top$ that models the pixel offset, with respect to a central point, between the leading and trailing edge polarity of the observed blob.
We expand the state model $x_k =(p_k, v_k, \theta_k, q_k, \lambda_k, \Delta_k)$ to include the offset parameter.
We add rotation dynamics with stochastic diffusion for the offset parameter to the state dynamics \eqref{eq:ODE}
\begin{align}
		\td \Delta & = J \Delta(t) \td t +  (Q^\Delta)^{\frac{1}{2}}\td w_\Delta,  
	\end{align}
	recalling $J$ from \eqref{eq:J} and $\Delta_k = \Delta(t_k)$.
	The offset is used to compensate the event position $\xi_k$ in the first pseudo measurement function
	\begin{align}
		H^{+}(x_k; \xi_k) := \Lambda_k^{-1}(\xi_k-\rho_k \Delta_k-p_k)
	\end{align}
	where $\rho_k$ is the event polarity at event timestamp $t_k$.
	Note that the polarity offset will converge to the corresponding direction as the velocity estimate if the blob is a light blob on a dark background and to the opposite direction to the velocity if it is a dark blob on a light background in order to balance the event polarities correctly.
	An analogous change is applied to the second pseudo measurement function \eqref{eq:zk}.

	\subsection{Prediction Step Integration}\label{sec:prediction_computation}
	Since events are triggered in microsecond time resolution, the continuous-time prediction step in Equation \eqref{eq:dynamics}-\eqref{eq: state covariance pred} are well-approximated by a first-order Euler integration scheme.
	For each timestamp $t_k$ drawn from the events associated with a specific blob, define the time interval
	\begin{align}
		\delta_k := t_{k} - t_{k-1}.
		\label{eq:delta_k}
	\end{align}
The predicted estimate of the prior state $\hat{x}_k^-$ and state covariance $\hat{\Sigma}^-_{k}$ are computed by
	\begin{subequations}
		\begin{align}
			\hat{x}^-_{k} & \approx (I_8 + \delta_k A_k)  \hat{x}_{k-1}, \\
			\hat{\Sigma}^-_{k} & \approx (I_8 + \delta_k A_k)  \hat{\Sigma}_{k-1}  (I_8 + \delta_k A_k)^\top + \delta_k Q_k.
		\end{align}
	\end{subequations}

\subsection{Computational Cost}

The AEB tracker includes two main steps: data association and the extended Kalman filter.	
The data association step has a linear event complexity of $\mathcal{O}(N)$ for $N$ input events, effectively passing only a subset of $M$ events to the extended Kalman filter, where $M \ll N$.
The extended Kalman filter operates with a time complexity of $\mathcal{O}(nM)$, where $n$ denotes the buffer length in~\eqref{eq:buffer1}-\eqref{eq:buffer2}.
In practice, the buffer length is typically set to $n < 10$, and it is worth noting that $N$ remains significantly larger than $nM$.
As a result, our algorithm maintains an overall linear event complexity of $\mathcal{O}(N)$.

Since the data association step only scales linearly with the event rate and this process significantly reduces the number of events used in the Extended Kalman filter, the algorithm is highly efficient for real-time implementation.
The efficiency and asynchronous processing of the AEB tracker make it well-suited for implementation on FPGA.
Such systems have the potential to be particularly advantageous for high-performance embedded robotic systems.


\section{Experimental Evaluation}\label{sec:experiments}

	\begin{figure*}[t!]
		\centering
		\begin{tabular}{c}
			\includegraphics[width=1\linewidth]{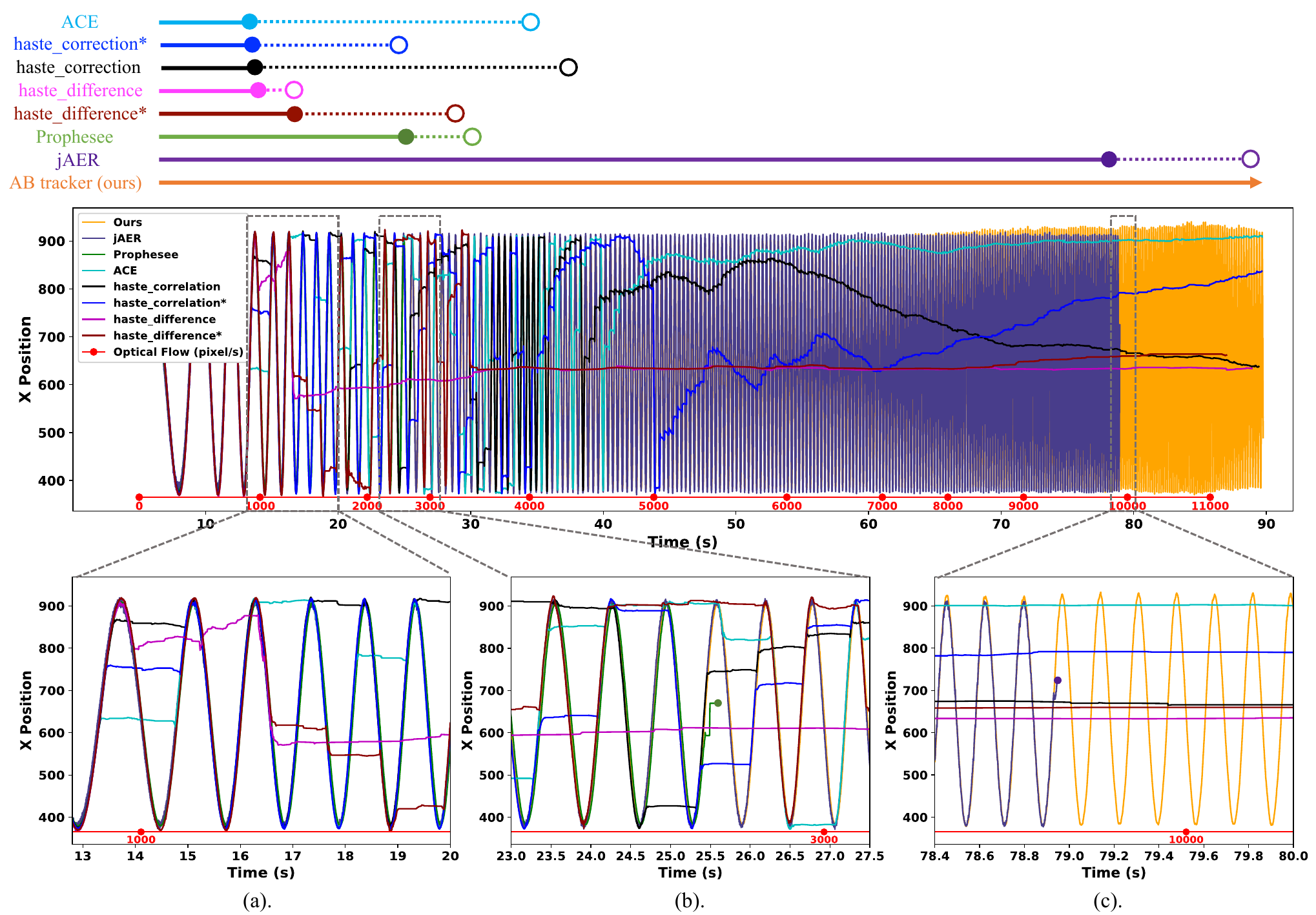}
		\end{tabular}
		\caption{
Example of evaluating the asynchronous event-based tracking methods ACE~\cite{Alzugaray18threedv}, HASTE~\cite{alzugaray2020haste}, Prophesee~\cite{Prophesee}, jAER~\cite{jAER} and our AEB tracker.
The target moves in a circular trajectory, increasing speed from $\sim$100 pixels/s to more than 11000 pixels/s over 90 seconds.
The main figure illustrates the $x$ position estimate generated by each
algorithm.
The lines above this figure show the speed range in which each algorithm operates robustly without losing track.
Dotted continuation of these lines indicates that the algorithm is able to
reinitialise but continues to fail with progressively shorter life times.
The solid dot terminations indicate the end-of-life of the target track while the open circle terminations to the dotted line indicates that future tracks last less than one cycle.
The sub-figures (a)-(c) provide zoomed-in detail in short time windows
around the key changes in behaviour.
Figure (a) shows the ACE~\cite{Alzugaray18threedv} and the four HASTE algorithms~\cite{alzugaray2020haste} losing track.
These algorithms initially can recapture track as the target circles and
recrosses the current state estimate.
Figure (b) shows the Prophesee~\cite{Prophesee} method losing track.
The Prophesee~\cite{Prophesee} deletes track hypothesis  that no longer correspond to an active track (indicated by the solid dot termination in the subplot).
A new ID could be selected to continue tracking, however, without prior knowledge about which of the possible targets correspond to the desired blob such a process cannot be considered a natural part of the tracking algorithm.
Figure (c) demonstrates the jAER~\cite{jAER} method losing target track.
Once again the algorithm provides a collection of alternative hypothesis for targets to track each of which can be tracked for a short time, however, the initial loss of track indicates the speed at which the algorithm becomes unreliable.
		}\label{fig:comparison-curve}
	\end{figure*}

This section is the first of two results sections.
In this section we evaluate the AEB tracker in a controlled laboratory environments to provide comparative results to state-of-the-art event-based corner and blob tracking algorithms.
Later, in Section~\ref{sec:Case_Studies} we provide some real-world case studies.
We provide two comparisons in this section; firstly tracking a spinning target to demonstrate the maximum optical flow that can be tracked, and secondly, tracking ego-motion of blobs for fast unstructured camera motion comparing tracking performance with and without IMU data.

\subsection{Indoor Dataset Collection}
\begin{figure}[t]
	\centering
	\setlength{\tabcolsep}{1.2pt}  
	\renewcommand{\arraystretch}{0.2}  
	\begin{tabular}{c c c}
		\includegraphics[width=0.32\linewidth]{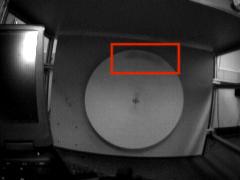} &
		\includegraphics[width=0.32\linewidth]{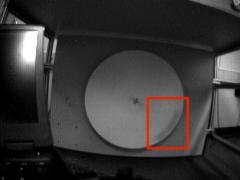} &
		\includegraphics[width=0.35\linewidth]{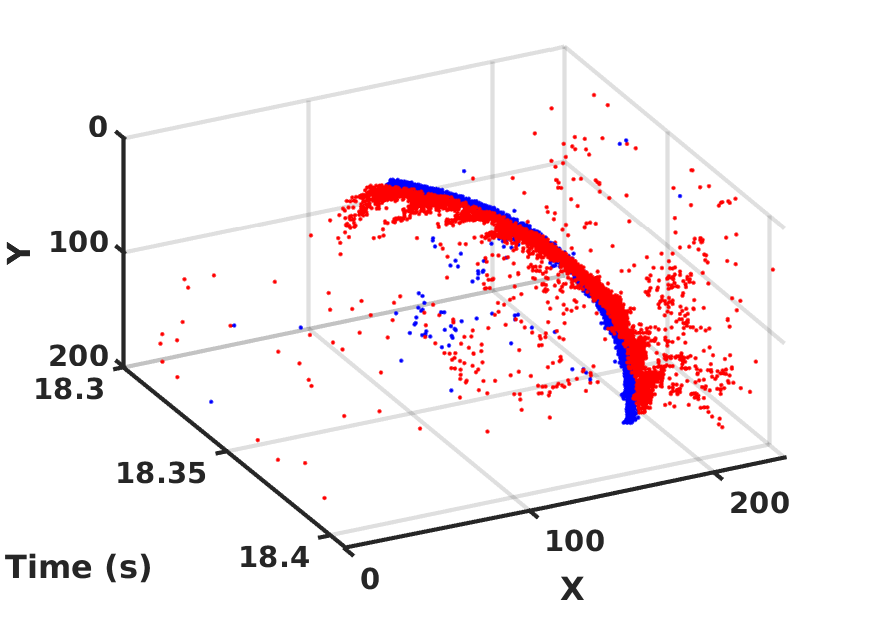} \\
		(a) & (b) & (c)
	\end{tabular}
	\caption{\label{fig:spin}
`Fast-Spinning' experiment.
The black target is attached to a spinning white disk and moving at $\sim$8000 pixels/s.
Figures (a)-(b) display two consecutive frames captured by a conventional frame-based camera operating at 20Hz.
The fast-moving target (highlighted by red boxes) is highly blurred in
a conventional image and discontinuous between the two consecutive images.
Figure (c) shows the event stream captured by the Prophesee Gen4 camera over the same period, demonstrating the microsecond time resolution necessary to track high-speed blobs.
}
\end{figure}

For the `Fast-Spinning' indoor experiment,
we record high-resolution events using a Prophesee Gen4 pure event camera (720$\times$1280 pixels).
We evaluate the tracking performance of different event-based corner and blob tracking algorithms at varying speeds.
To achieve this, we have constructed a spinning disk test-bed (Figure \ref{fig:spin}) with adjustable spin rates, and we use a black square on a white disk as the target to provide corners for the corner detection algorithms to operate.

For the `Fast Moving Camera with Ego-motion' experiment, we record data using a hybrid event-frame DAVIS 240C camera (180$\times$240 pixels) that outputs synchronised events, reference frames and IMU data.
Note that we do not record frame data because the hybrid DAVIS cameras usually generate a large number of noisy events at the shutter time of each frame when operating in hybrid mode \cite{wang2021stereo}.

\subsection{Fast-Spinning Data}

In this section, we evaluate our method against the most relevant state-of-the-art object-tracking algorithms that use only events as input, operate efficiently (preferably asynchronously), and require no training process.
The comparisons include ACE from~\cite{Alzugaray18threedv}, four HASTE methods from~\cite{alzugaray2020haste}, the generic object tracking method from the industrial Prophesee software~\cite{Prophesee} and the Rectangular Cluster Tracker (RCT) function from the jAER software~\cite{jAER}.
In addition, we broaden our scope by comparing with event-based corner detection methods, including the popular asynchronous corner tracking algorithm Arc* \cite{Alzugaray18ral} and the widely used benchmark method EOF~\cite{Zhu17icra}.
Note that we limit our comparisons to event-based methods due to the extreme nature of our targeted scenarios in this paper.
Frame-based cameras lack the capability to capture information for tracking fast motion or in poor lighting due to motion blur and jumps in blob location between frames (Fig.~\ref{fig:spin}).
All algorithms (where velocity is estimated) use a constant velocity model,
that is, linear trajectories for the blobs.
We do not provide algorithms with prior knowledge that the target trajectory is circular.
This mismatch makes the dataset challenging at sufficiently high optical flow.

\begin{figure}[t]
	\begin{tabular}{c}
		\includegraphics[width=0.9\linewidth]{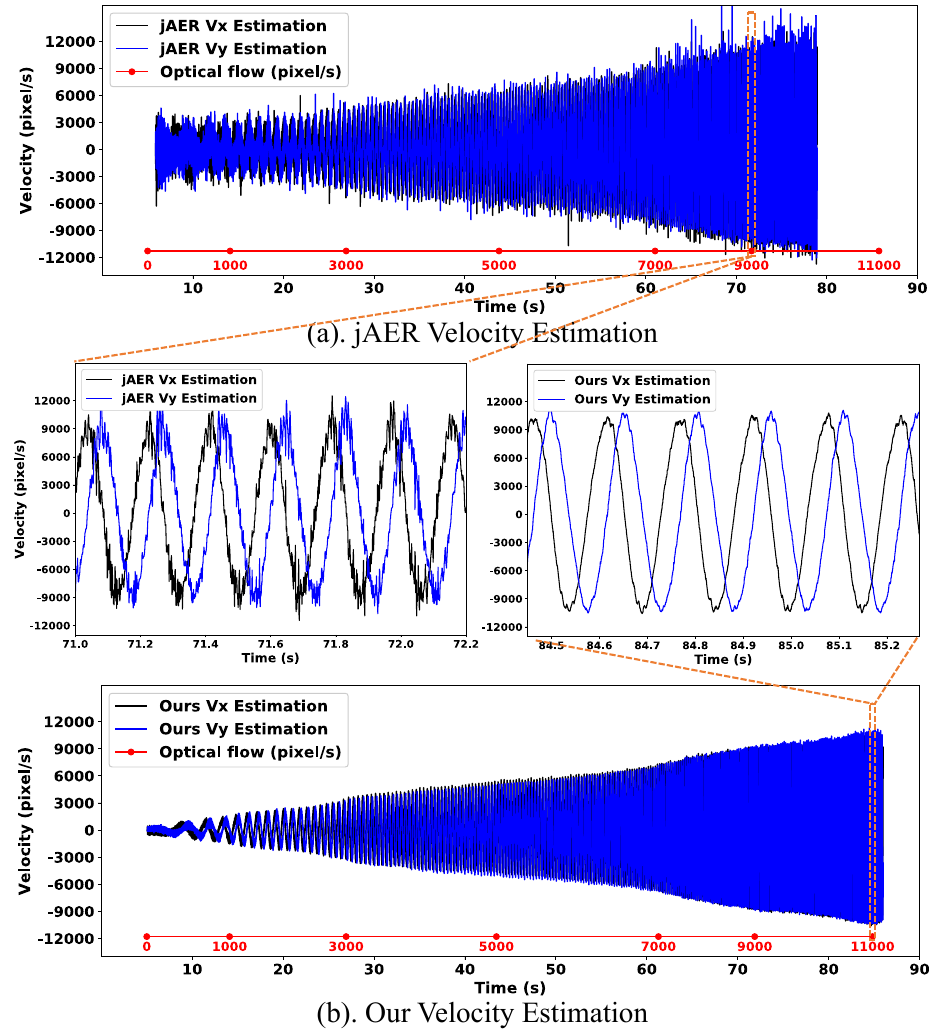}
	\end{tabular}
	\caption{\label{fig:speed esti}
			Example of evaluating the velocity estimation performance of jAER~\cite{jAER} and our AEB tracker in sub-figure (a) and (b).
			The target spins from $\sim$100 pixels/s to more than 11000 pixels/s.
			The reference optical flow of the tracked target is marked in red.
			The zoomed-in velocity estimate during two short time windows around 9000 and 11000 (pixels/s) are presented in the middle.
			The diagrams show that the velocity estimate of our AEB tracker closely aligns with the referenced velocity even at very high-speed and exhibits less noise than jAER~\cite{jAER}.
	}
\end{figure}

\begin{table}
	\centering
	\begin{tabular}{|c|c|}
		\hline
		Algorithms & Max Speed (pixels/s) \\
		\hline\hline
		ACE~\cite{Alzugaray18threedv} & \phantom{$>$} 914 \phantom{$>$}\\
		haste\_correction~\cite{alzugaray2020haste}  & \phantom{$>$} 939 \phantom{$>$}\\
		haste\_correction*~\cite{alzugaray2020haste}  & \phantom{$>$} 925 \phantom{$>$}\\
		haste\_difference~\cite{alzugaray2020haste}  & \phantom{$>$} 992 \phantom{$>$} \\
		haste\_difference*~\cite{alzugaray2020haste}  & \phantom{$>$} 1476 \phantom{$>$}
		\\
		Prophesee~\cite{Prophesee}  & \phantom{$>$} 2620 \phantom{$>$}
		\\
		jAER~\cite{jAER} & \phantom{$>$} 9830 \phantom{$>$}
		\\
		AB\_tracker (ours)  & $>$ 11320 \phantom{$>$} \\
		\hline
	\end{tabular}
	\vspace{0.1cm}
	\caption{
		Quantitative comparison results for asynchronous event blob tracking algorithms on the spinning dot data.
	}
	\label{tab:results}
\end{table}

\begin{figure}[]
	\begin{tabular}{c}
		\includegraphics[width=0.97\linewidth]{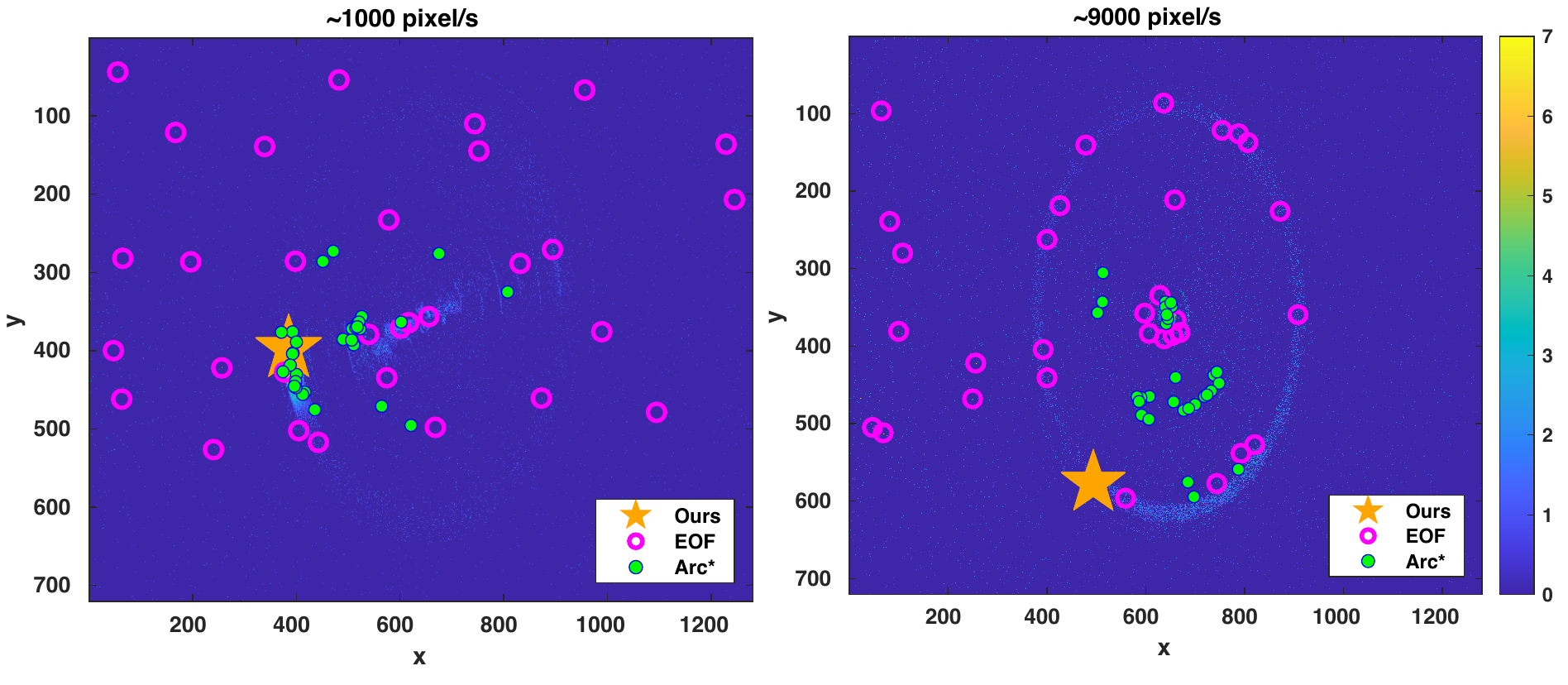}
	\end{tabular}
	\caption{\label{fig:eof}
Comparison of the proposed algorithm with two event-based corner detection algorithms EOF~\cite{Zhu17icra} and Arc* \cite{Alzugaray18ral}.
Corners detected are shown as purple circles (EOF~\cite{Zhu17icra}) or green dots (Arc* \cite{Alzugaray18ral}).
The background image is coloured (colour bar on right of figure) to show accumulated events per pixel over a short duration.
	}
\end{figure}

The main panel of Figure~\ref{fig:comparison-curve} plots the $x$-component of the blob position for each of the different algorithms during an experiment where the optical flow increases from $\sim$100 pixels/s to more than 11000 pixels/s over 90 seconds.
The graphic above the main panel plots the period of time for which each algorithm provides reliable tracking of the blob.
The plotted line indicates the period up to which the algorithm loses track the first time.
The solid dot terminations indicate the end-of-life of the target track while the open circle terminations to the dotted line indicates that future tracks last less than one cycle.
The bottom sub-figures highlight zooms of short windows of data to demonstrate key behaviour of the algorithms.
The first window, at $\sim$1000 pixels/s, captures the moment when the ACE~\cite{Alzugaray18threedv} and the four HASTE algorithms~\cite{alzugaray2020haste} begin to lose track.
These algorithms have enduring state estimates and since the blob is travelling in a circular trajectory they often recapture the blob at a later moment as it recrosses the location of the state estimate.
This intermittent tracking persists for some time but tracking failures become more regular and eventually the estimates drift from the circular route of the blob and tracking is lost.
We also note that ACE~\cite{Alzugaray18threedv} and HASTE~\cite{alzugaray2020haste} depend heavily on good identification of the feature template and this is only reliably possible at low speeds.
The second figure captures the moment that the Prophesee~\cite{Prophesee} method loses track, while the third window captures the moment that the jAER~\cite{jAER} loses track.
The Prophesee~\cite{Prophesee} was run with default parameters while the jAER~\cite{jAER} was hand-tuned working with the author (jAER~\cite{jAER}) to achieve the best performance on the test dataset possible.
Both these algorithms delete target hypotheses when they lose track.
This is indicated with a terminal point on the track in the zoomed windows.
Both these algorithms also include detection routines and initialise new targets, which (if they happen to be the correct blob) are able to maintain track for a few additional cycles.
However, the first tracking failure indicates the onset of non-robustness in the tracking. Both algorithms regularly lose track for optical flow higher than that indicated in the graphic above the main plot.
Our proposed method continues to track stably up to the physical limitations of the experimental platform ($>$ 6 rev/s).
Table~\ref{tab:results} presents the maximum optical flow that each algorithm could reliably maintain track.

In addition to the position estimation, jAER~\cite{jAER} and our proposed method also estimate the target velocity.
In Figure~\ref{fig:speed esti}, we plot the velocity estimates $V_x = \dot{x}$ and $V_y = \dot{y}$ of the target by the jAER~\cite{jAER} and our method.
The figures illustrate the linear increase in peak velocity of the oscillation to around 11000 pixels/s in both the $x$ and $y$ directions.
The two zoomed-in plots show the detailed behaviour of the velocity estimates of both algorithms for $\sim$9000 pixels/s and $\sim$11000 pixels/s and demonstrate the stability of our algorithm in comparison to the much noisier estimates provided by the jAER algorithm~\cite{jAER}.

Apart from the blob tracking methods, we have also provided qualitative comparison studies with event-based corner detection methods, including the popular asynchronous corner tracking algorithm Arc* \cite{Alzugaray18ral} and the widely used benchmark method EOF~\cite{Zhu17icra} at $\sim$1000 pixels/s and $\sim$9000 pixels/s is shown in Figure \ref{fig:eof}.
We mark the corners detected by EOF~\cite{Zhu17icra} by green filled dots, Arc* \cite{Alzugaray18ral} by magenta circles and the position estimated by our proposed method by an orange star.
The two compared algorithms mostly detect corners around the target at low speed but fail at high speed and mainly detect the centre of the spinning test bed.
Neither algorithm is able to provide reliable target tracking at high optical flow for this simple scenario.

\begin{figure}[t]
	\centering
	\setlength{\tabcolsep}{0.6pt}  
	\renewcommand{\arraystretch}{0.4}  
	\begin{tabular}{lcc}
		\raisebox{1cm}{$t_0$ \hspace{0.3cm}} &
		\includegraphics[width=0.41\linewidth]{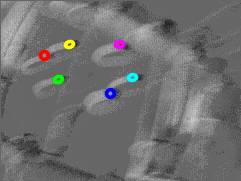} &
		\includegraphics[width=0.41\linewidth]{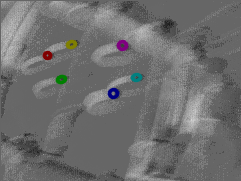} \\
		\raisebox{1cm}{$t_1$} &
		\includegraphics[width=0.41\linewidth]{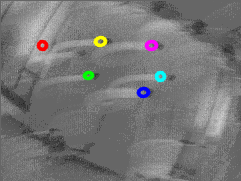} &
		\includegraphics[width=0.41\linewidth]{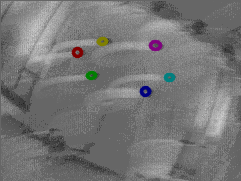} \\
		& (a) no IMU & (b) with IMU (b)
	\end{tabular}
	\caption{\label{fig:imu-dots}
Example of tracking six black targets under rapid camera motion.
Freeze frames at timestamps $t_0$ and $t_1$ are shown with a time difference of $0.025$ second.
(a) Tracking without modelling camera ego-motion.
(b) Tracking with modelling camera ego-motion using IMU data.
The tracked targets are highlighted by coloured circles in (a), while in (b), targets are marked with the same colours but appear darker.
The shadows in the background are windowing effects from event accumulation and reflect the rapid ego-motion of the image blobs.
Notably, the targets in (a) are slightly behind the true blob positions compared to (b).
Moreover, in (a) without camera ego-motion compensation, the red target loses track at $t_1$.
	}
\end{figure}

\begin{figure}[t]
	\centering
	\begin{tabular}{c}
		\includegraphics[width=0.7\linewidth]{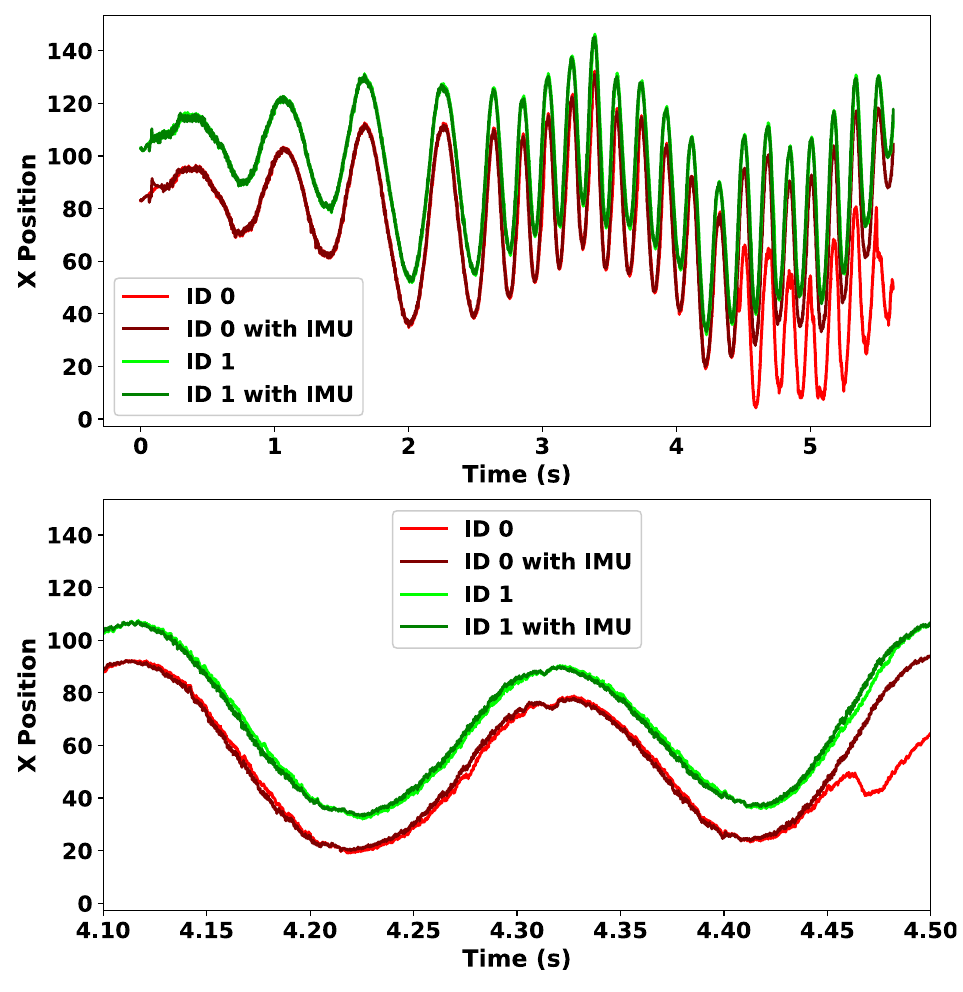}
	\end{tabular}
	\caption{\label{fig:imu-curve}
		The trajectory of the two targets marked in red and green in Figure \ref{fig:imu-dots}.
			The upper plot shows the full trajectory of $x$ position estimations, while the lower plot is a zoomed-in view around time $t = 4.30$ s.
			When camera ego-motion is not modelled using IMU data (brighter colour), the trajectories consistently lag slightly behind their corresponding trajectories in darker colour.
The red target (no IMU) loses track at time $t = 4.45$ s.
}
\end{figure}

\subsection{Fast Moving Camera with Ego-motion}
This section evaluates the effectiveness of compensating for camera ego-motion using IMU data in our tracker.
The experiment involved tracking the black shapes on a white background in an image where the camera was rapidly shaken by hand.
The dataset is similar to, and we used the same shapes image, as the popular dataset proposed by \cite{Mueggler17ijrr}, but with more aggressive camera motions, and all targets remain in the camera field-of-view.
Maximum optical flow rates are $\sim$1000 pixels/s with abrupt changes in direction.

In Figures \ref{fig:imu-dots} and \ref{fig:imu-curve}, we compare the tracking trajectory of our AEB tracker with and without including camera ego-motion using IMU data.
Brighter colours represent trajectories where no IMU data is available, while darker colours correspond to the case where gyroscope measurements from a camera-mounted IMU are available and used to predict ego-motion in the filter as outlined in Section~\ref{sec:ABtracker}.
Figure~\ref{fig:imu-curve} demonstrates that the `no IMU' trajectories lag slightly behind those of the ego-motion compensated model.
This effect is also visible in Fig.~\ref{fig:imu-dots} where the position estimates of the `no IMU' filter are lagging the latest event data slightly more than the estimates obtained using IMU to compensate ego-motion.
Compensating camera ego-motion also reduces overshoot when the camera motion changes direction at high speed (Fig.~\ref{fig:imu-curve}).
The combination of these effects makes the `no IMU' case slightly less robust and we were able to induce a tracking failure when the camera was shaken at nearly 6Hz with optical flow rates between $\pm$1000 pixels/s.
The authors note that although including camera ego-motion is clearly beneficial, the performance of the algorithm without ego-motion compensation is better than expected and is sufficient for many real-world applications.

\begin{figure}[t!]
	\centering
	\begin{tabular}{c}
		\includegraphics[width=0.9\linewidth]{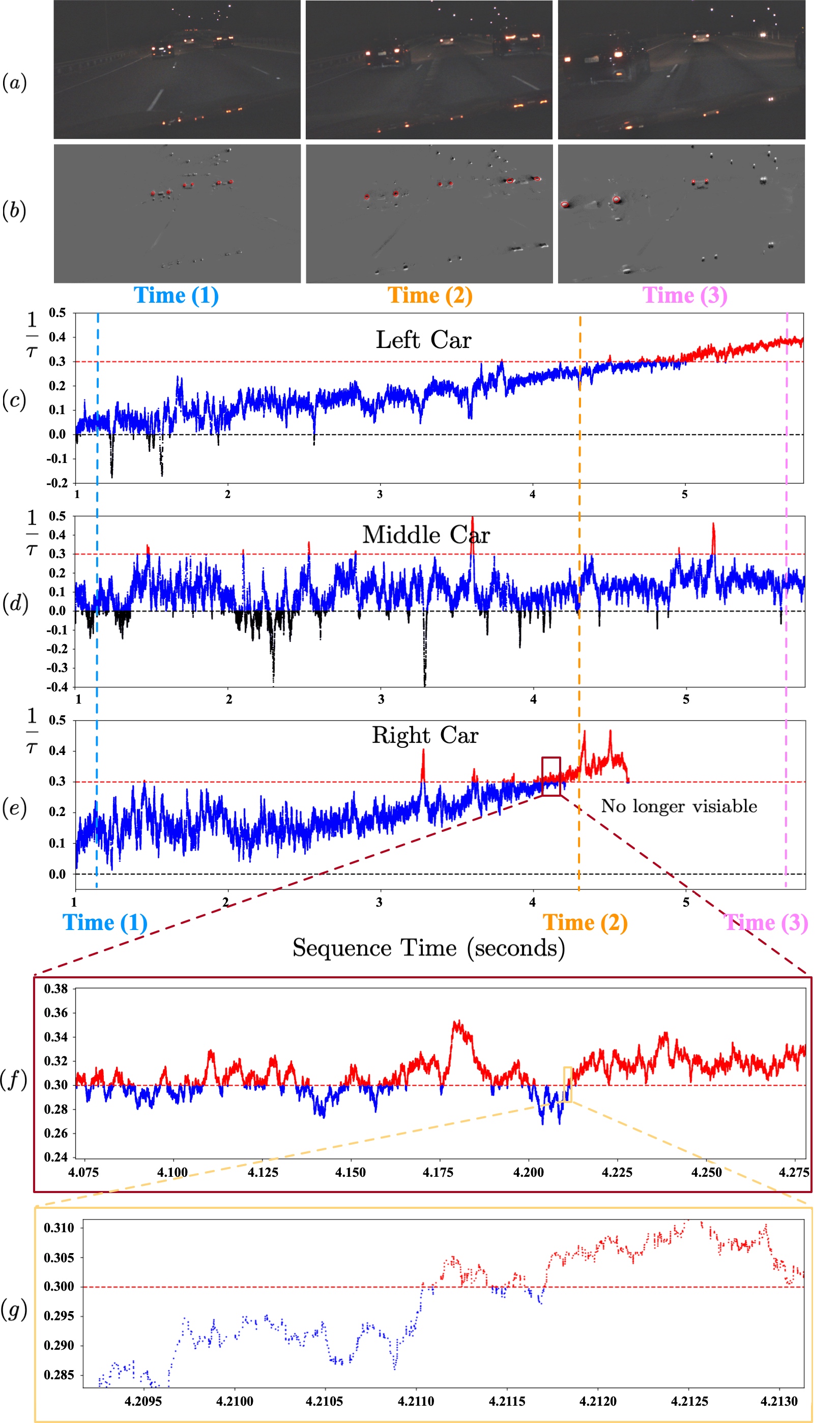}
	\end{tabular}
	\caption{\label{fig:ttc}
Multi-vehicle tail lights tracking and time-to-contact (TTC) experiments.
Three cars are visible beneath street lights on a multi-lane road.
The cars tail lights are reflected in the bonnet of the experimental vehicle and only the actual tail lights are tracked.
(a): Reference RGB frames.
(b): Event reconstructions obtained using the event-based high-pass filter~\cite{Scheerlinck18accv}. Tracked objects are marked with red ellipses.
(c)-(e): Estimated (inverse) time-to-contact from the three front cars.
Blue represents that two cars approaching and black means they are diverging.
Red represents that two cars are approaching faster than the safety threshold.
The timestamps of the three reference images are marked in three dotted lines.
(f): The zoomed-in view of the red box section in (e).
(g): The further zoomed-in view of the yellow box section in (f), demonstrating the proposed filter achieves more than 100kHz visual TTC estimation.
See more details in the supplementary video.
}\end{figure}

\section{Real-World Case Study}
\label{sec:Case_Studies}
In this section, we showcase our AEB tracker in two challenging real-world scenarios: tracking automotive tail light at night and tracking a quickly moving drone in complex environments, with applications to estimating time-to-contact and range estimation.

\subsection{Outdoor Data Collection}
\begin{figure}[t]
	\centering
	\begin{tabular}{c}
		\includegraphics[width=0.9\linewidth]{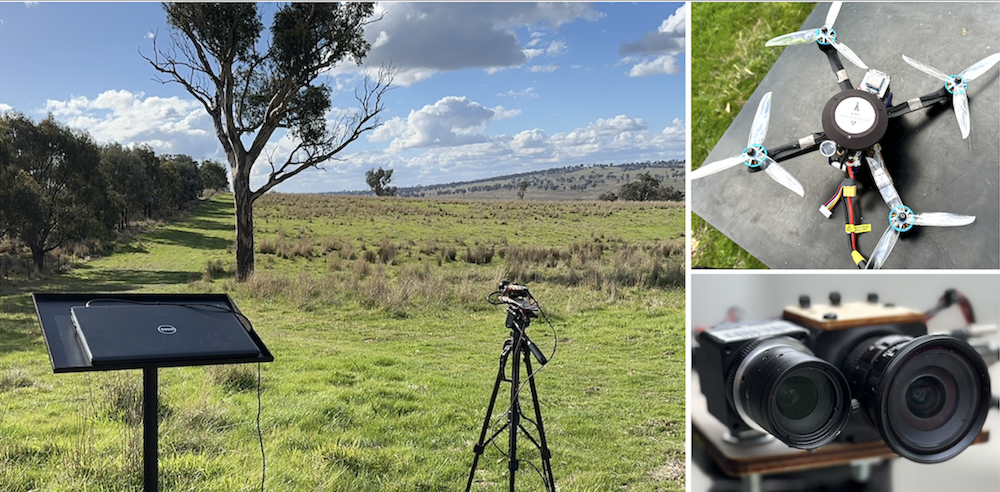}
	\end{tabular}
	\caption{\label{fig:stereo rig}
		Left: outdoor flying quadrotor experiments. Right: Our quadrotor with an RTK-GPS and our stereo event-frame camera. }\end{figure}

Apart from recording event data using a Prophesee camera, we also use a FLIR RGB frame camera (Chameleon3USB3, 2048$\times$1536 pixels) to capture high quality image frames for the real-world case study.
The two cameras were placed side-by-side and time-synchronised by an external trigger (see Figure~\ref{fig:stereo rig}).
The RGB camera frames were re-projected to the event camera image plane using stereo camera calibration.
The re-projection mismatch due to the disparity between the two cameras is minor for far-away outdoor scenes.
Both datasets include high-speed targets and challenging lighting conditions, allowing a comprehensive evaluation of tracking and shape estimation, as well as for downstream tasks such as time-to-contact estimation and range estimation.
For the flying quadrotor dataset, we used a small quadrotor equipped with a Real-Time Kinematic (RTK) positioning GPS (see Figure~\ref {fig:stereo rig}).

\subsection{Night Driving Experiment}

In this experiment, the event camera system was mounted on a tripod (held down by hand in the foot well) inside a car driving on a public road at night.
The proposed algorithm is used to track the tail lights of other vehicles, and we demonstrate its performance by using the output states to estimate time-to-contact (TTC) at more than 100kHz.

\begin{figure*}[t!]
	\centering
	\begin{tabular}{c}
		\includegraphics[width=0.9\linewidth]{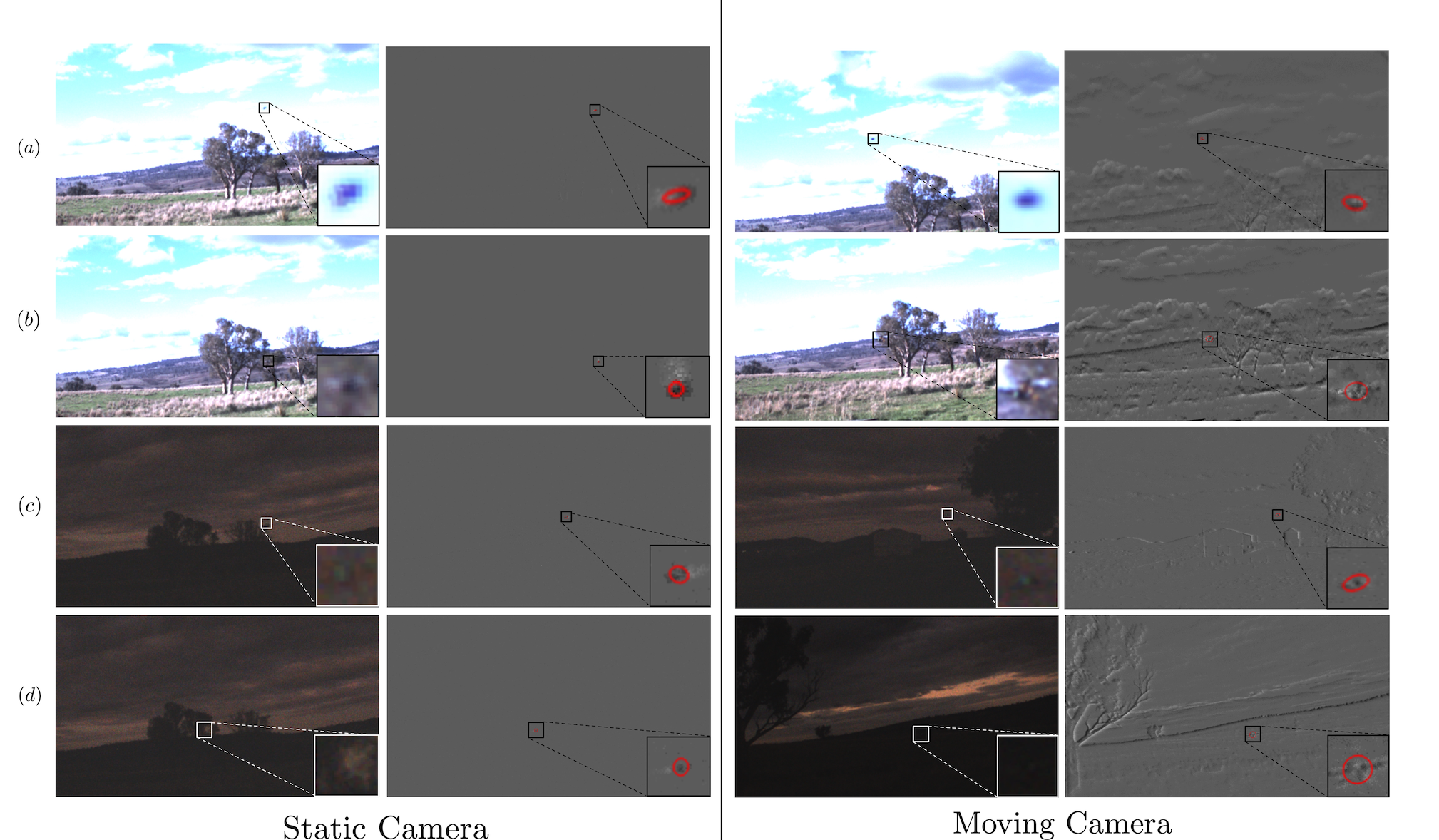}
	\end{tabular}
	\caption{\label{fig:drone}
		Example of tracking a quadrotor in different lighting conditions and camera motions using only events.
		Grayscale images are displayed only for reference.
		The estimated position, size and orientation of the drone are illustrated by red ellipses.
		Refer to the supplementary video for the tracking performance.
}\end{figure*}

\subsubsection{Time-To-Contact}
Time-to-contact (TTC) defines the estimated time before a collision occurs between two objects, typically a vehicle and an obstacle or another vehicle.
We estimate TTC by computing the rate of variation of the pixel distance between the left and right tail lights on the front cars in the image \cite{negre2008real, gonner2009vehicle}.
Our filter provides high-rate estimates of the position and velocity of the event blobs associated with the left and right tail lights on each car at time $t_k$, denoted as $\hat{p}^L_k$, $\hat{p}^R_k$, $\hat{v}^L_k$, and $\hat{v}^R_k$, respectively.
The visual distance between the tail lights is defined to be
\begin{align}
	s & :=  \| \hat{p}^L_k - \hat{p}^R_k \|,
\end{align}
in pixels.
The relative velocity between the left and right tail lights is defined to be
the relative blob velocity in the direction of displacement (in pixels/s),
\begin{align}
	v & := (\hat{v}^L_k - \hat{v}^R_k)^\top \frac{ \hat{p}^L_k - \hat{p}^R_k}{\|\hat{p}^L_k - \hat{p}^R_k\|}.
\end{align}
The TTC, measured in seconds, is the ratio of the visual distance $s$ between the left and right tail lights to the relative visual velocity $v$ between the lights on the same car in pixels \cite{negre2008real, gonner2009vehicle}
\begin{align}
	\tau(t_k) :=  \frac{s}{v}.
\end{align}
The inverse TTC
\begin{align}
\frac{1}{\tau(t_k)} :=   \frac{(\hat{v}^L_k - \hat{v}^R_k)^\top (\hat{p}^L_k - \hat{p}^R_k)}{\|\hat{p}^L_k - \hat{p}^R_k\|^2}.
\end{align}
is most commonly used as a cue for obstacle avoidance since it is a bounded signal and setting warning thresholds is straightforward as shown in Figure~\ref{fig:ttc}.

\subsubsection{Experiments}
In Figure~\ref{fig:ttc}, we present an example of tracking the tail lights of three cars at night, travelling with an average speed of approximately 90~km/h.
The figure displays the reference images captured by an RGB frame-based camera in (a) and the event reconstructions (obtained using the event high-pass filter~\cite{Scheerlinck18accv}) in (b) across three different timestamps.
Only the actual tail lights of the cars were tracked (marked with red ellipses on images (b)) and not the reflections of the tail lights visible in the bonnet of the car.
By choosing the bias of the Prophesee Gen4 camera carefully, we are able to preserve events associated with the high-frequency LED lights and reduce background ego-motion events leading to high signal-to-noise event data.
This trivialises the data association and enables robust, accurate, high-band width performance of our event blob tracker.
Conversely, the RGB reference frames are mostly dark and blurry and obtaining dynamic information on relative motion in the image from this data would be difficult.

We estimate the inverse time-to-contact (TTC) $\frac{1}{\tau}$ of the left, middle, and right car separately and plot them in Figure~\ref{fig:ttc}c)-e).
The timestamps corresponding to the RGB and event images in the top rows are marked in the plots below; timestamps (1), (2), and (3) are distinguished by a blue, orange, and pink dashed line, respectively.
A positive $\frac{1}{\tau}$ indicates that the distance to a given car is decreasing, while a negative $\frac{1}{\tau}$ indicates that the distance to a given car is increasing (and consequently no collision can occur).
For illustrative purposes, we have chosen to indicate a safety threshold at $\frac{1}{\tau} = 0.3$ when the following car is around 3.3 seconds from collision with the leading car if the velocities were to be kept constant.
Values exceeding this are shown in red in Figure~\ref{fig:ttc}c)-g).

Figure~\ref{fig:ttc}d) indicates the experimental vehicle maintaining a relatively constant TTC to the middle car in the same lane, consistent with safe driving margins.
Figure~\ref{fig:ttc}e) shows the inverse TTC $\frac{1}{\tau}$ of the right car exceeds the safety threshold at approximately 4.3 seconds, shortly before we overtake it at 4.6 seconds.
No TTC data is shown in Figure~\ref{fig:ttc}e) after 4.6 seconds since the car is not visible.
In Figure~\ref{fig:ttc}c), the inverse TTC of the left car exceeds the safety threshold around 5 seconds.
See the RGB frames and event reconstructions in Figure~\ref{fig:ttc}a)-b) for reference.

Irregularities in the road surface induce bouncing and swaying motion of the vehicle frame that, since the camera is held firmly to the dashboard of the car, lead to horizontal linear velocity of the camera and results in rapid changes in TTC.
There are several clear short peaks in \ref{fig:ttc}d), both positive and negative, due to small bumps and depressions in the road surface.
For example, the brief 50-millisecond black (negative) peak that appears around 3.3 seconds in Figure~\ref{fig:ttc}d) along with the red peak in Figure~\ref{fig:ttc}e) would indicate that the left front wheel has encountered a bump inducing the vehicle to pitch backward and sway to the right.
The motion can be appreciated in the supplementary video for the full sequence, although since the entire bump sequence occurs in a period of around 50ms the actual motion of the car is correspondingly small.

Figure~\ref{fig:ttc}f) provides a zoomed-in view of the red box section in Figure~\ref{fig:ttc}e) of 200 milliseconds in length, and Figure~\ref{fig:ttc}g) provides a further zoomed-in view of the yellow box section within subplot (f) of 4 milliseconds in length.
These plots demonstrate the exceptionally high temporal resolution of our tracker, processing updates at more than 100kHz.
This remarkable speed minimises latency for vital decision-making processes such as braking and acceleration, particularly in emergency situations.
It clearly surpasses the capabilities of frame-based TTC algorithms, which are limited by the frequency of incoming RGB frames (typically around 40Hz), or even pseudo-frame event based algorithms (requiring around 1K to 10K events per pseudo-frame) that would operate in the order of 10-100Hz.
This performance, for a single event camera at night and without any additional active sensors like Lidar or Radar, demonstrates the potential of applying event cameras, and the proposed AEB tracker,  to problems in autonomous driving.

\subsection{Flying Quadrotor Experiment}
We demonstrate the performance of the proposed AEB tracker by tracking a flying quadrotor under various camera motions and lighting conditions shown in Figure~\ref{fig:drone}.
The average speed of the drone in the image is around 150 pixels/s, with some fast motions, such as fast flips, reaching speeds of 400 pixels/s.
Figure~\ref{fig:drone}a)-b) demonstrates the tracking performance in well-lit conditions, while data shown in Figure~\ref{fig:drone}c)-d) demonstrate the performance in poorly lit conditions.
The RGB images and event reconstruction images are provided for reference only and the filter is implemented on pure asynchronous event data.
The red ellipses superimposed on the event reconstruction images illustrate the position, size, and orientation of the quadrotor as estimated by our AEB tracker.
The minor axis of the ellipse aligns with the drone's heading direction.
Please also refer to the supplementary video to see the tracking performance.

The first two columns in Figure~\ref{fig:drone} show the reference frames and event reconstruction recorded with a stationary camera.
In these conditions, the majority of the events are triggered by the movement of the quadrotor since it is the only object in the scene that changes brightness or motion.
Therefore, the event reconstruction frames mainly consist of a gray background with an event blob triggered by the target.
With good lighting conditions and a simple background, the target is easily distinguishable in both frames and event data (see Figure~\ref{fig:drone}a).
The advantage of our proposed algorithm over frame-based algorithms becomes more apparent in low-light conditions or environments with complex backgrounds.
As shown in Figure~\ref{fig:drone}b), the complex background in the frames makes tracking using RGB frames extremely challenging when the quadrotor crosses trees.
In contrast, our  AEB tracker can still track the quadrotor successfully.
In the poorly lit sequences in Figure~\ref{fig:drone}c)-d), the RGB frames only capture a faint green light emitted by an LED on the quadrotor, making tracking nearly impossible.
On the other hand, the event camera can still capture unique event blobs of the quadrotor even in low light or in front of a complex background, providing sufficient information for our  AEB tracker to achieve reliable tracking performance even in these challenging scenarios.

In the third column in Figure~\ref{fig:drone}, when the camera is also moving, the background triggers many events, posing a challenge for tracking.
The dynamic nearest neighbour data association approach (cf. Section \ref{sec:data-assoc}) selectively feeds events to the filter based on their proximity to the centre of the target and estimation of the target size.
Only events within a dynamic threshold from the target's centre are considered, effectively rejecting the vast majority of background noise.
Although overall the tracking performance is excellent, the present algorithm can still lose track if substantial ego-motion of the camera occurs just as the target crosses a high contrast feature in the scene.
This is particularly the case if the camera ego-motion tracks the target, since this reduces the optical flow of the target and increases the optical flow of the scene.
In such situations, a scene feature blob can capture the filter state and draw it away from the target.
Interestingly, this scenario is almost always related to the rotational ego-motion of the camera, since such motion generates the majority of optical flow in a scene.
The corresponding filter velocity of the incorrect blob track could then be computed \textit{a-priori} from the angular velocity of the camera, and this prior information could be exploited to reject false background tracks. 
Implementation of this concept is beyond the scope of the present paper but is relevant in the context of the future potential of the algorithm.

\subsubsection{Qualitative Tracking Performance}
\begin{figure}[t!]
	\centering
	\begin{tabular}{c}
		\includegraphics[width=0.9\linewidth]{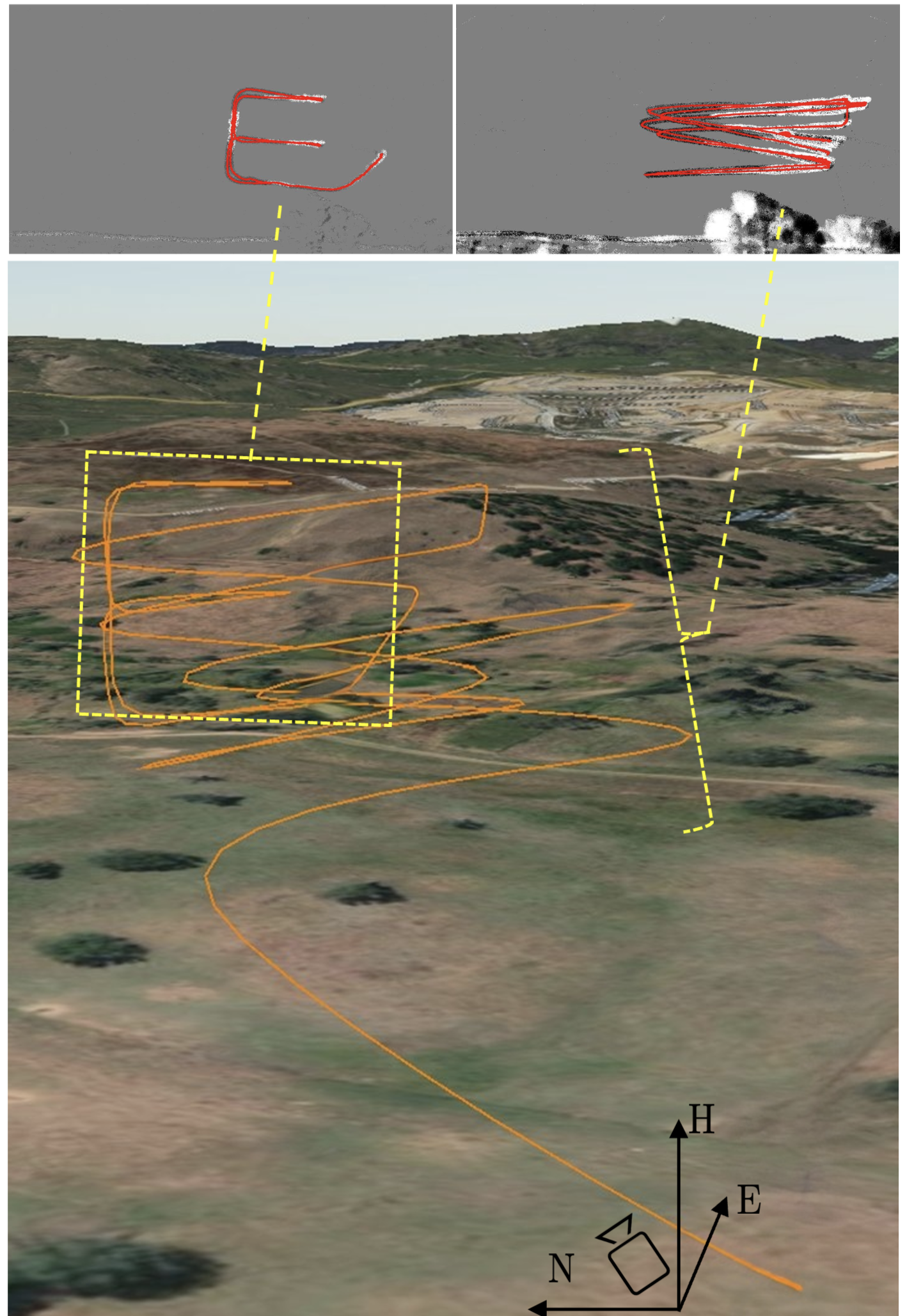}
	\end{tabular}
	\caption{\label{fig:trajectory}
	Example of the estimated trajectory by our AEB tracker and the 3D GPS	trajectory in a 3D plot.
	The first row shows the estimated trajectory (marked by red dots) of the tracked quadrotor overlaid on the event reconstruction frames.
	The second row shows the 3D GPS data of the full trajectory. The camera position and East-North-Up direction are marked at the bottom.
}\end{figure}

To provide a qualitative measure of tracking performance we have overlaid the estimated trajectory of the tracked quadrotor on event reconstructions (obtained using the event-based high-pass filter~\cite{Scheerlinck18accv}) during aerobatic manoeuvres.
In Figure~\ref{fig:trajectory}, red dots represent the continuous-time trajectory estimated by our AEB tracker while the background frames show the reconstructed intensity image from events.
The close alignment between the red estimated trajectory and the observed event trajectory shows that our tracker accurately tracks the trajectory of the target quadrotor.
The quadrotor was also equipped with RTK-GPS and the full 3D trajectory is shown as an orange trajectory in the perspective image in Figure~\ref{fig:trajectory}.
The corresponding zigzag and `E' sub-trajectories are indicated in yellow.
The camera position and East-North-Up direction are marked at the bottom right of the figure.

\subsubsection{Range Estimation}
The shape parameters estimated in the proposed filter allow us to derive an estimate of range for a target with known size.
The target shape parameters $(\lambda^1_k, \lambda^2_k)$ encode the visual second-order moment of the event blob.
Since the quadrotor is disc shaped, the maximum of the visual second-order moment is visually related to the diameter of the quadrotor.
By triangulation, the range can be estimated by
\begin{align}
 \text{Range} &= \frac{a f}{\textrm{max}(\lambda^1_k, \lambda^2_k)},
\end{align}
where $a$ is the actual diameter of the target in meters and $f$ is the camera focal length.
Note that we estimate the real-world target range based on the object's real-world size and the camera's focal length as prior knowledge, assuming the target size is constant.

\begin{figure}[t!]
	\centering
	\begin{tabular}{c}
		\includegraphics[width=1\linewidth]{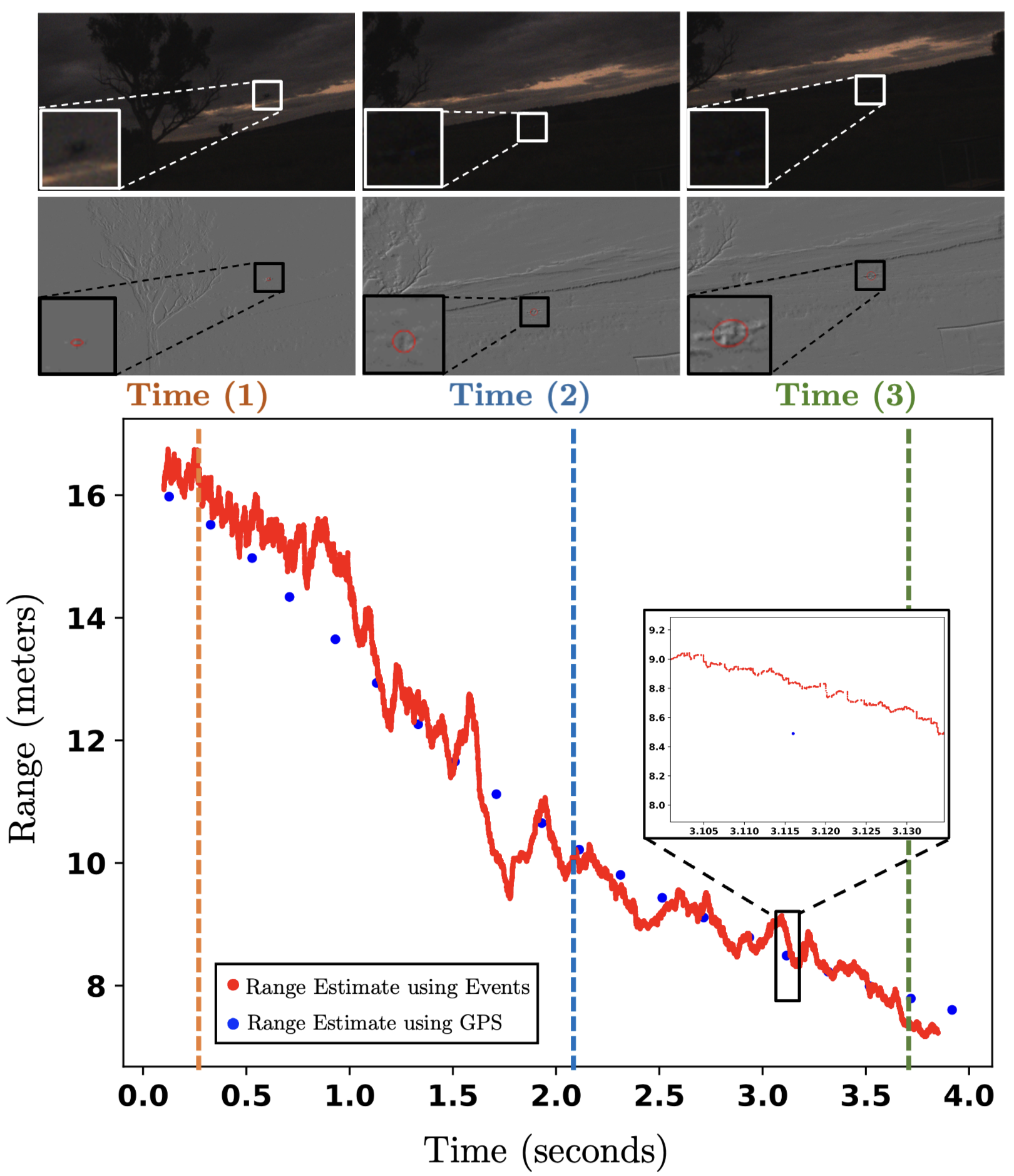}
	\end{tabular}
	\caption{\label{fig:range}
		Range estimation of an approaching quadrotor in a low-light environment using events and GPS data.
		The first and second rows show the reference RGB images and event reconstructions.
		The estimated position and shape using events are represented by red ellipses on event reconstructions.
		The third row shows the continuous-time range estimate along with a zoom-in plot illustrating a short period.
		The estimated range aligns with the range estimation using GPS data and achieves more than a 50kHz update.
}\end{figure}

We provide an example of range estimation for a quadrotor as it approaches the camera location in low-light conditions from a distance of approximately 17 meters to a distance of less than 8 meters in Figure~\ref{fig:range}.
The image data is shown at three time-stamps each indicated by a different colour while the detailed range estimation data is shown in red in the following graph.
The reference data points (in blue) are computed from RTK-GPS log from the vehicle compared to a known GPS location of the camera.
Note that the RTK-GPS is only available at 5Hz and our AEB tracker achieves more than a 50kHz update.

The quadrotor is almost impossible to visually identify or track using the dark RGB images in Figure~\ref{fig:range}.
This is particularly the case when it flies in front of the dark mountains in the background.
The precision of the proposed range information is strongly coupled with the high filter update rates that allow for temporal smoothing of the inferred shape parameters.
In particular, even though the virtual measurement function based on the chi-squared statistic does involve temporal averaging, the time window on which this statistic is computed is so short (due to the high filter update rate) compared to the vehicle dynamics that the proposed range estimate is remarkably reliable.

\section{Limitations and Future Works}

The proposed algorithm uses a simple and computationally efficient data association method that is suitable for many general tasks.
However, this simplicity limits its application in tasks with crossing targets and large amounts of extraneous events generated by textured background and ego-motion of the camera.
Future research avenues include the design of a more robust data association algorithm for general tasks and the creation of specific data association methods tailored to specialised tasks that are robust to ego-motion.
A second avenue for further work is to develop a robust detection method
method and building a complete detection and tracking system to complement the spatio-temporal Gaussian likelihood model for event blob target tracking.
Finally, it is of interest to extend the approach track general spatio-temporal features rather than just blobs.
			
\section{Conclusion}
This paper presents a novel asynchronous event blob tracker for tracking event blobs in real time. The proposed filter achieves accurate tracking and blob shape estimation even under low-light environments and high-speed motion.
It fully exploits the advantages of events by processing each event and updating the filter state asynchronously, achieving high temporal resolution state estimation (over 50kHz for the experiments undertaken).
The algorithm outperforms state-of-the-art event tracking algorithms in laboratory based experimental testing, achieving tracking rates in excess of 11000 pixels/s.
The high-rate, robust filter output directly enables downstream tasks such as range estimation and time-to-contact in autonomous driving, demonstrating the practical potential of the algorithm in a wide range of computer vision and robotics applications that require precise tracking and size estimation of fast-moving targets.

\begin{figure*}[t]
	\centering
	\setlength{\tabcolsep}{0.5pt}  
	\renewcommand{\arraystretch}{0.3}  
	\begin{tabular}{lccccc}
		\raisebox{1cm}{$(a)$} &
		\includegraphics[width=0.2\linewidth]{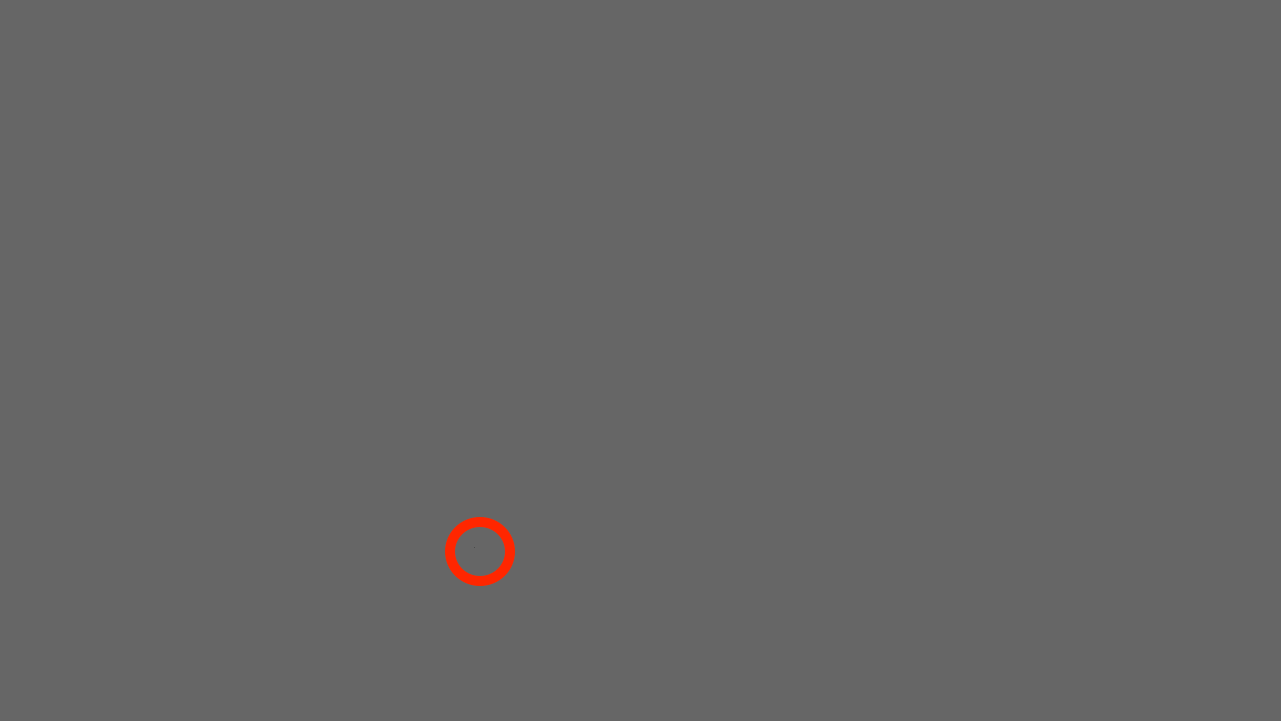} &
		\includegraphics[width=0.2\linewidth]{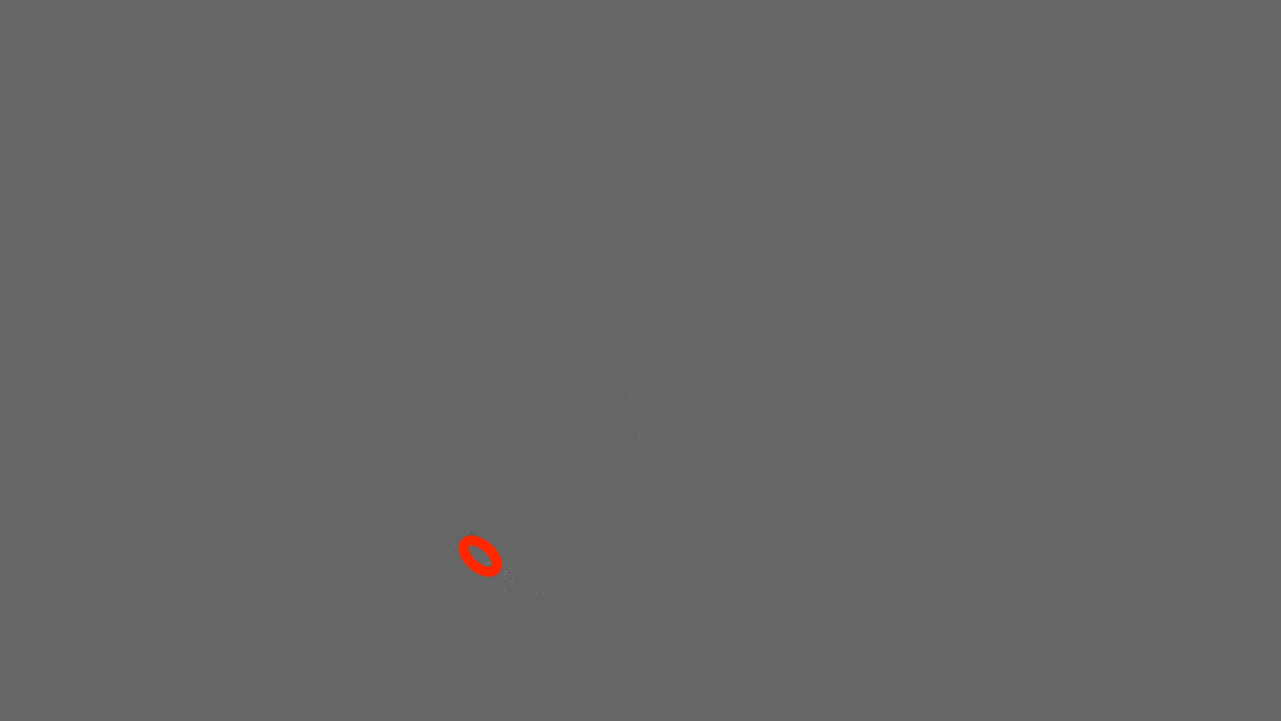} &
		\includegraphics[width=0.2\linewidth]{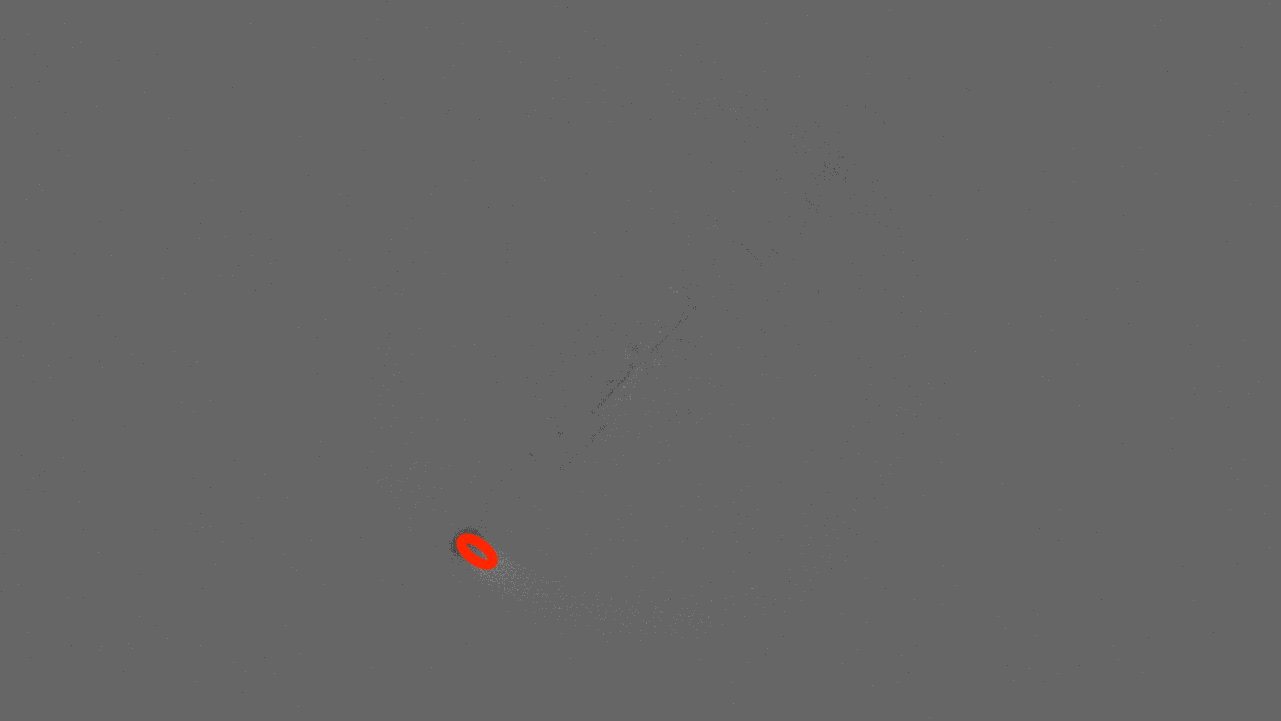} &
		\includegraphics[width=0.2\linewidth]{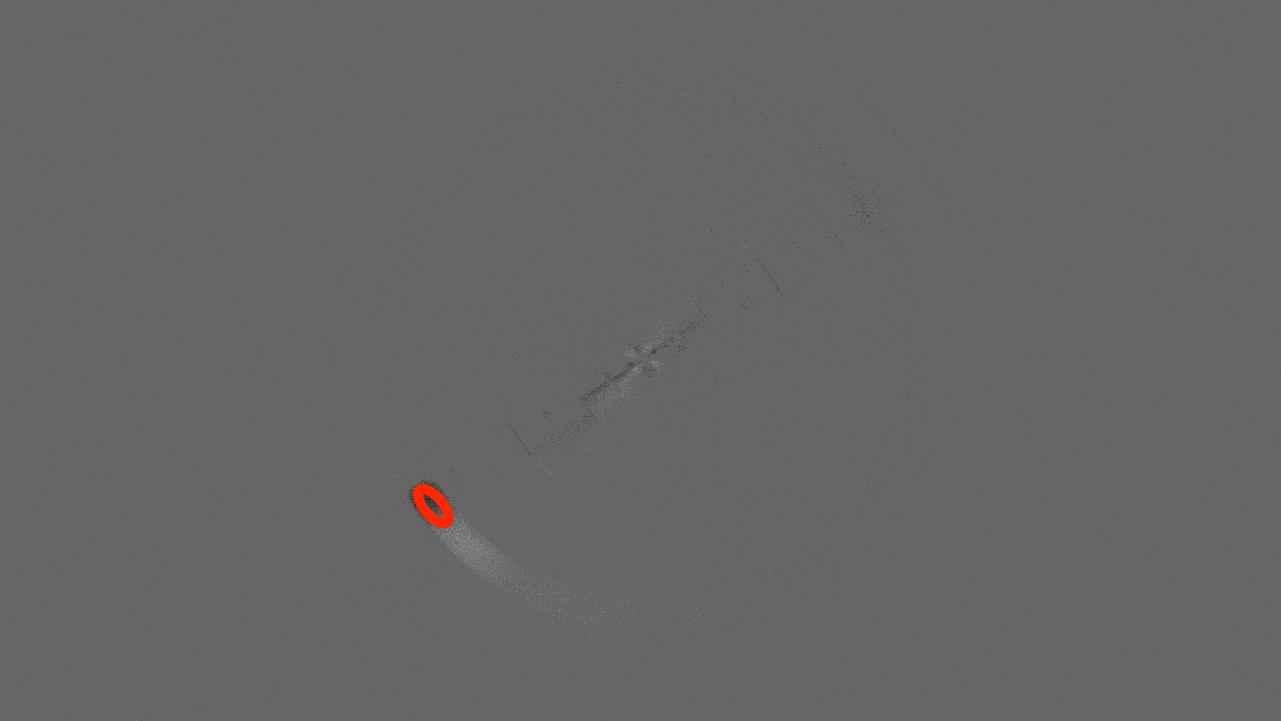} &
		\includegraphics[width=0.2\linewidth]{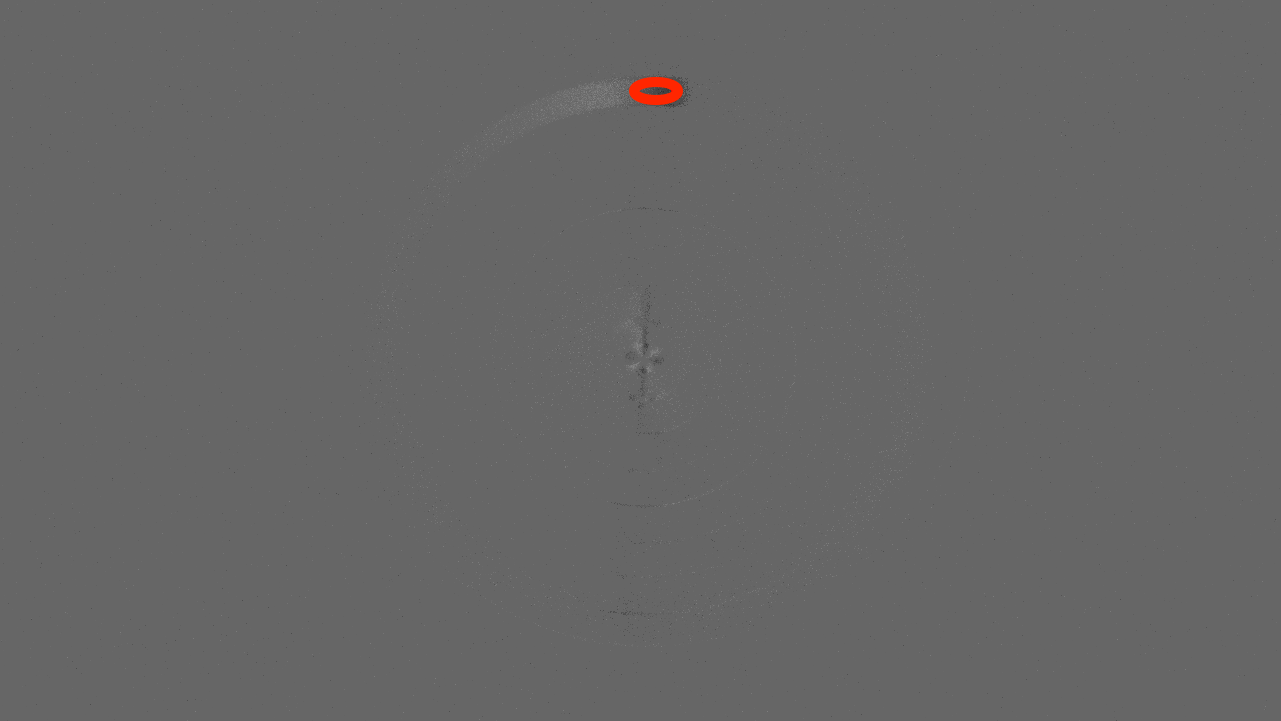} \\
		\raisebox{1cm}{$(b)$} &
		\includegraphics[width=0.2\linewidth]{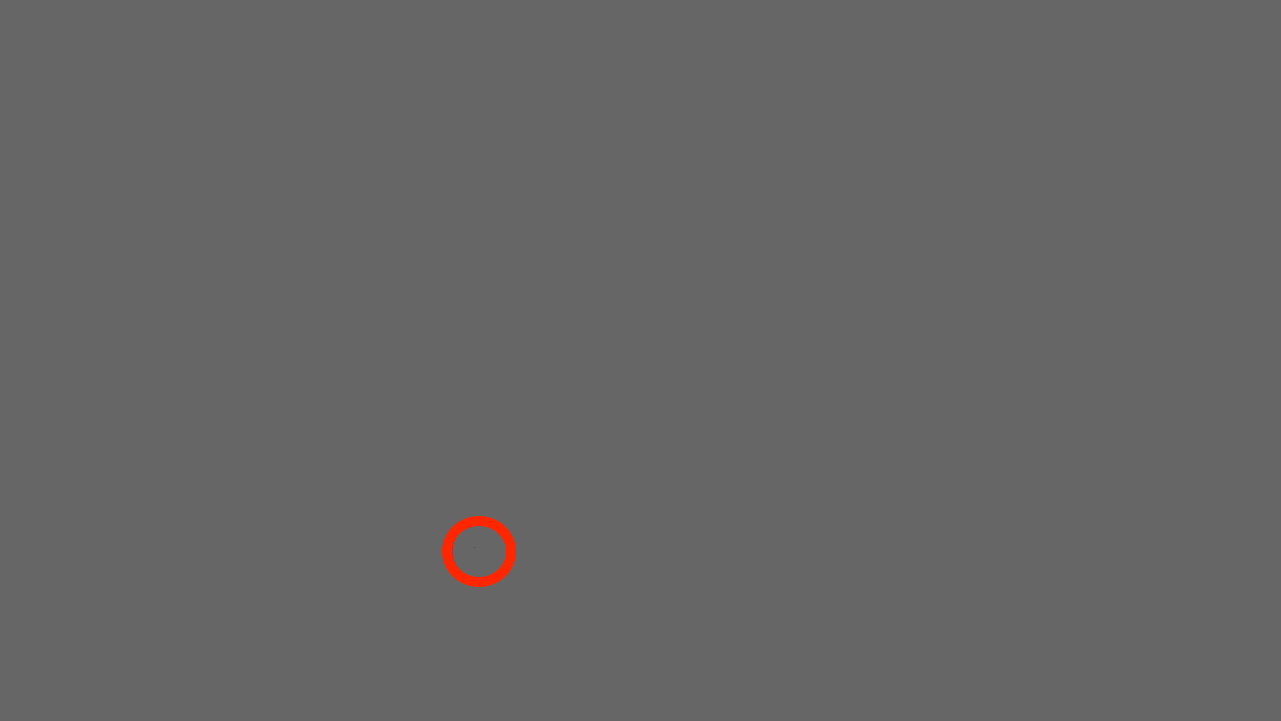} &
		\includegraphics[width=0.2\linewidth]{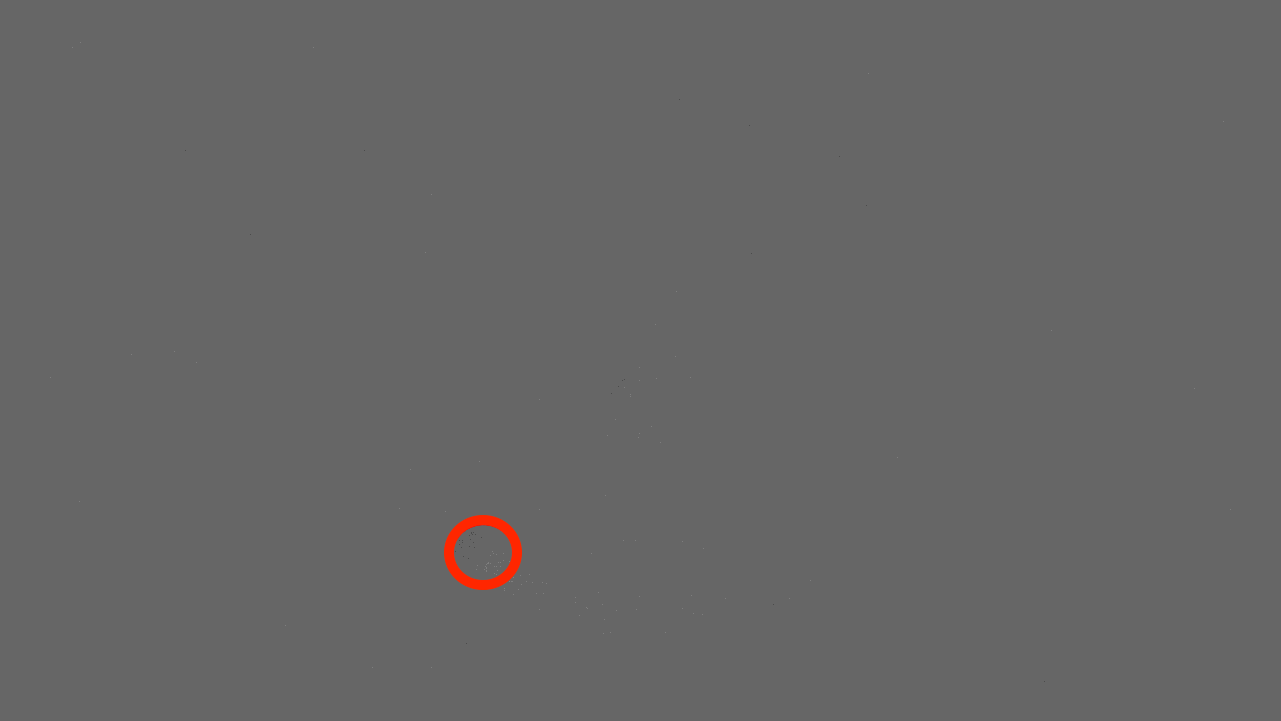} &
		\includegraphics[width=0.2\linewidth]{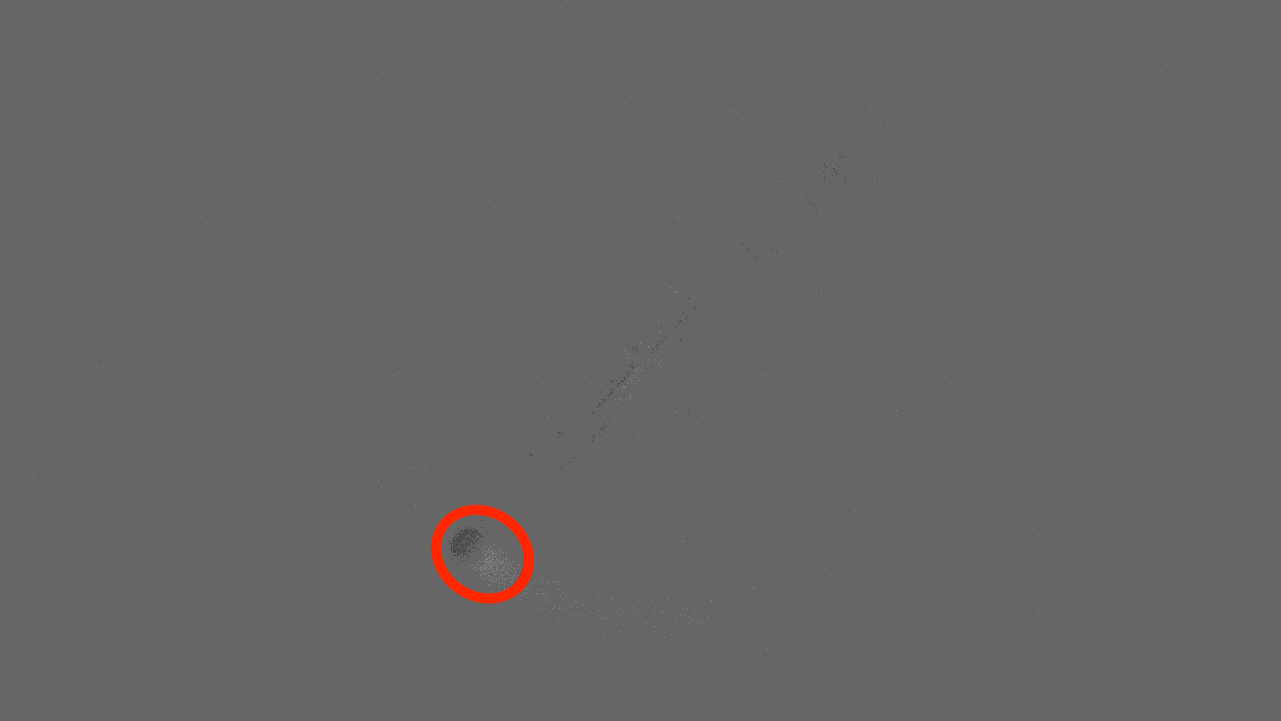} &
		\includegraphics[width=0.2\linewidth]{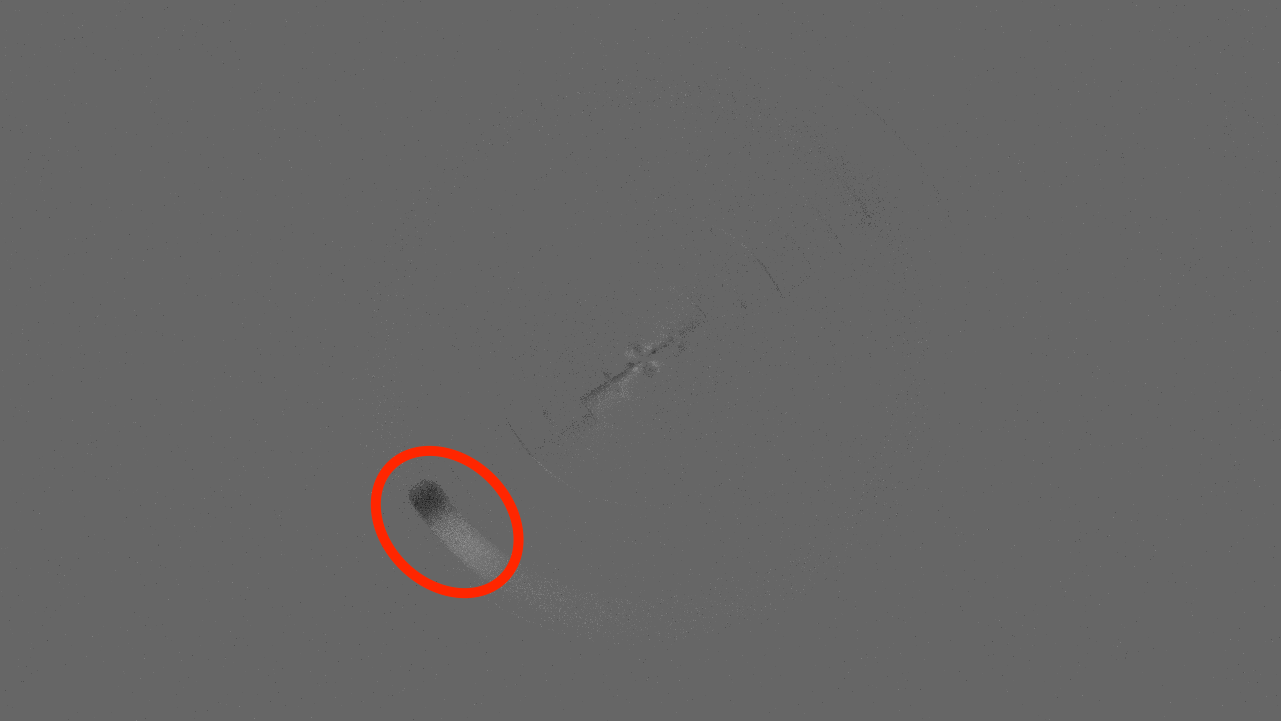} &
		\includegraphics[width=0.2\linewidth]{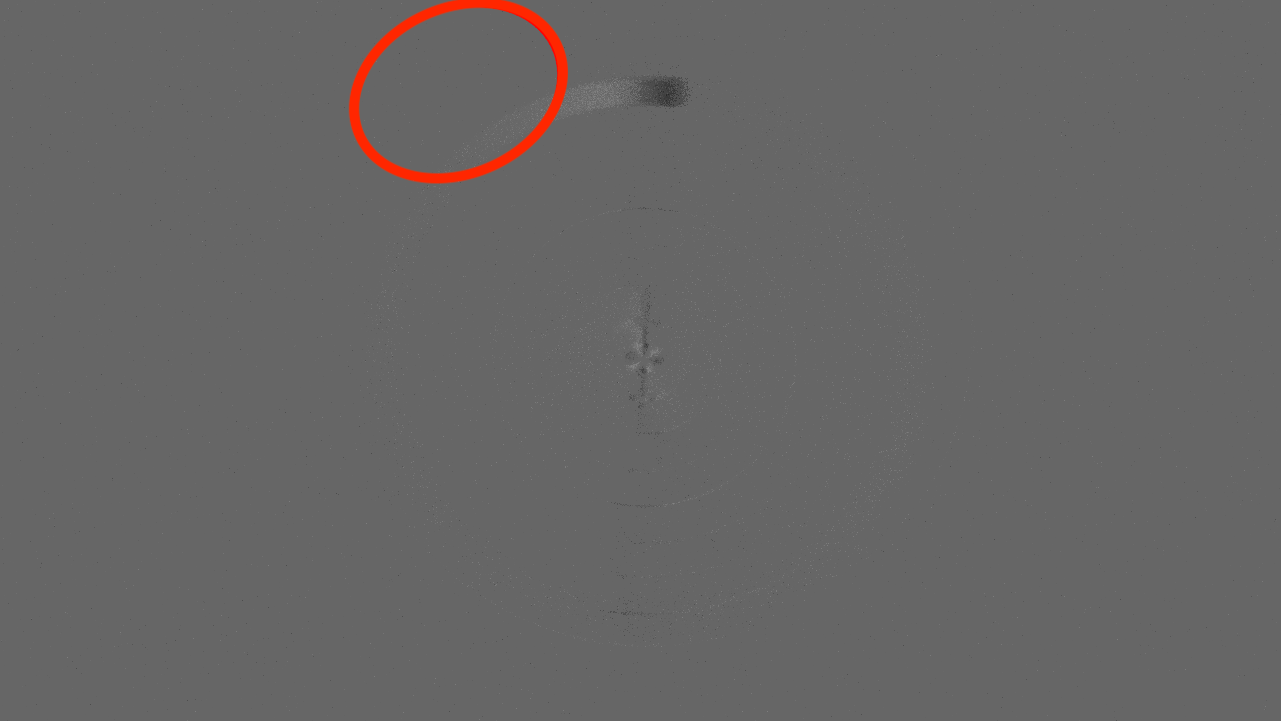} \\
		& $t_0$ & $t_0 + 0.2 (ms)$ & $t_0 + 4 (ms)$ & $t_0 + 40 (ms)$ & $t_0 + 360 (ms)$
	\end{tabular}
	\caption{\label{fig:ablation}
			Ablation study evaluating the effectiveness of the second measurement function $z_k$ in shape parameter estimation.
			The tracking performance is demonstrated at different timestamps, comparing the use of the combined measurement function $m_k$ in (a), with only the first measurement function $y_k$ in (b). The red ellipses represent the estimated target position and size.
	}
\end{figure*}

\section{Acknowledgement}
The authors would like to thank Professor Tobias Delbruck of ETH Zurich for his guidance in using and tuning parameters within the jAER software~\cite{jAER}, along with his insightful suggestions to this paper.
We also sincerely thank Dr. Andrew Tridgell and Ms. Michelle Rossouw from the Ardupilot Community for their invaluable help collecting and processing the quadrotor dataset.
We thank Dr. Iain Guilliard and Mr. Angus Apps from the Australian National University for their contributions in modularising the codebase to enhance accessibility.

\section{Appendix}

\subsection{Partial Derivatives of Observation Model} \label{sec:app-Observation Model}
The partial derivatives of $H(x_k; \xi_k)$ with respect to $p_{k}$, $v_{k}$ and $\lambda_k$ are given by
\begin{align}
\begin{split}
D_{p_{k}}H(x_{k}; \xi_{k}) &= -\Lambda_k^{-1}, \\
D_{v_{k}}H(x_{k}; \xi_{k}) &= 0, \\
D_{\lambda_k}H(x_{k}; \xi_{k}) &= -\mathcal{R} (\theta_k) \diag[ \mu^2 \mathcal{R} (-\theta_k) (\xi_k - p_k)], \\
D_{\theta_k}H(x_{k}; \xi_{k}) &= \mathcal{R} (\theta_k) [\Omega \mu - \mu \Omega ] \mathcal{R} (-\theta_k) (\xi_k - p_k), \\
D_{q_{k}}H(x_{k}; \xi_{k}) &= 0, \\
\end{split}
\end{align}
The partial derivatives of $G(x_k; \hat{P}_k^-, \Xi_k)$
 with respect to $p_{k}$, $v_{k}$ and $\lambda_k$ are given by
\begin{align}
\begin{split}
D_{p_{k}}G(x_k; \hat{P}_k^-, \Xi_k)&=  0, \\
D_{v_{k}}G(x_k; \hat{P}_k^-, \Xi_k) &= 0, \\
D_{\lambda_{k}}G(x_k; \hat{P}_k^-, \Xi_k)
&= - \frac{2}{1+\beta} \sum_{j=1}^{n}  \zeta,
\\
D_{\theta_{k}}G(x_k; \hat{P}_k^-, \Xi_k) &= 0, \\
D_{q_{k}}G(x_k; \hat{P}_k^-, \Xi_k) &= 0,
\end{split}
\end{align}
where
\begin{align}
	\begin{split}
		\mu &:= \diag^{-1}[\lambda_k^1, \lambda_k^2], \\
		\Omega &= (0, -1; 1, 0),\\
		\zeta &:=\chi_{k-j}^\top  \mathcal{R} (\hat{\theta}_{k-j}^-)  \diag[\mu^2 \mathcal{R} (-\hat{\theta}_{k-j}^-) (\xi_{k-j} - \hat{p}^-_{k-j})]. \\
	\end{split}
\end{align}

\subsection{Ablation Study} \label{Ablation Study}

In this section we present an ablation study to evaluate the effectiveness of our second measurement function $z_k$ in the combined measurement function $m_k$ (\ref{eq:mk}) and experimentally demonstrate that adding $z_k$ enhances the observability of the shape parameter $\lambda$, as discussed in Section \ref{sec:pseudo-measurements}.

We use the `Fast-Spinning' dataset for this comparison.
Figure \ref{fig:ablation} illustrates the size estimation and tracking performance of using the combined measurement function $m_k$ in (a) and using only the first measurement function $y_k$ in (b).
The background intensity is generated using  high-pass filter \cite{Scheerlinck18accv} for display only.
The estimated target position and size are marked by red ellipses.

We initialise the target position and size at timestamp $t_0$. After 0.2 milliseconds, the ellipse computed by the combined measurement function in (a) quickly shrinks to the corresponding target size, while the ellipse of using the single measurement $y_k$ in (b) remains almost unchanged.
Subsequently, the ellipse in (a) further shrinks to the actual target size and the filter tracks the moving target accurately. However, the ellipse in (b) using only the first measurement function $y_k$, tracks the target but demonstrates a gradual increase in estimated size and eventually loses track.

\bibliographystyle{IEEEtran}
\bibliography{main}

\section{Biography Section}

\begin{IEEEbiography}[{\includegraphics[width=1in,height=1.25in,clip,keepaspectratio]{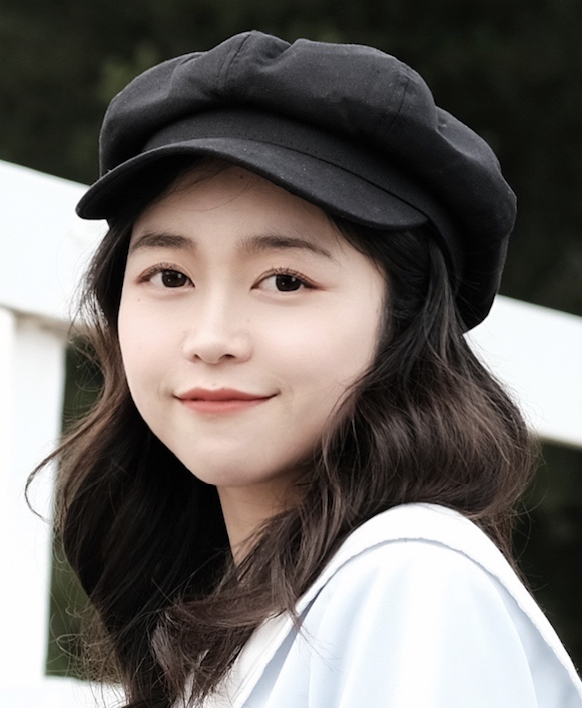}}]{Ziwei Wang}
	is a Ph.D. student in the Systems Theory and Robotics (STR) group at the College of Engineering and Computer Science, Australian National University (ANU), Canberra, Australia.
	She received her B.Eng degree from ANU (Mechatronics) in 2019.
	Her interests include event-based vision, asynchronous image processing and robotic applications.
	She is an IEEE student member.
\end{IEEEbiography}

\begin{IEEEbiography}[{\includegraphics[width=1in,height=1.25in,clip,keepaspectratio]{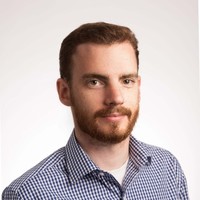}}]{Timothy L. Molloy} (Member, IEEE) was born in Emerald, Australia. He received the BE and PhD degrees from the Queensland University of Technology (QUT) in 2010 and 2015, respectively. He is currently a Senior Lecturer in the School of Engineering at the Australian National University (ANU). Prior to joining ANU, he was an Advance Queensland Research Fellow at QUT (2017-2019), and a Research Fellow at the University of Melbourne (2020-2022). His interests include signal processing and information theory for robotics and control.
Dr. Molloy is the recipient of a QUT University Medal, a QUT Outstanding Doctoral Thesis Award, a 2018 Boeing Wirraway Team Award, and an Advance Queensland Early-Career Fellowship.\end{IEEEbiography}

\begin{IEEEbiography}[{\includegraphics[width=1in,height=1.25in,clip,keepaspectratio]{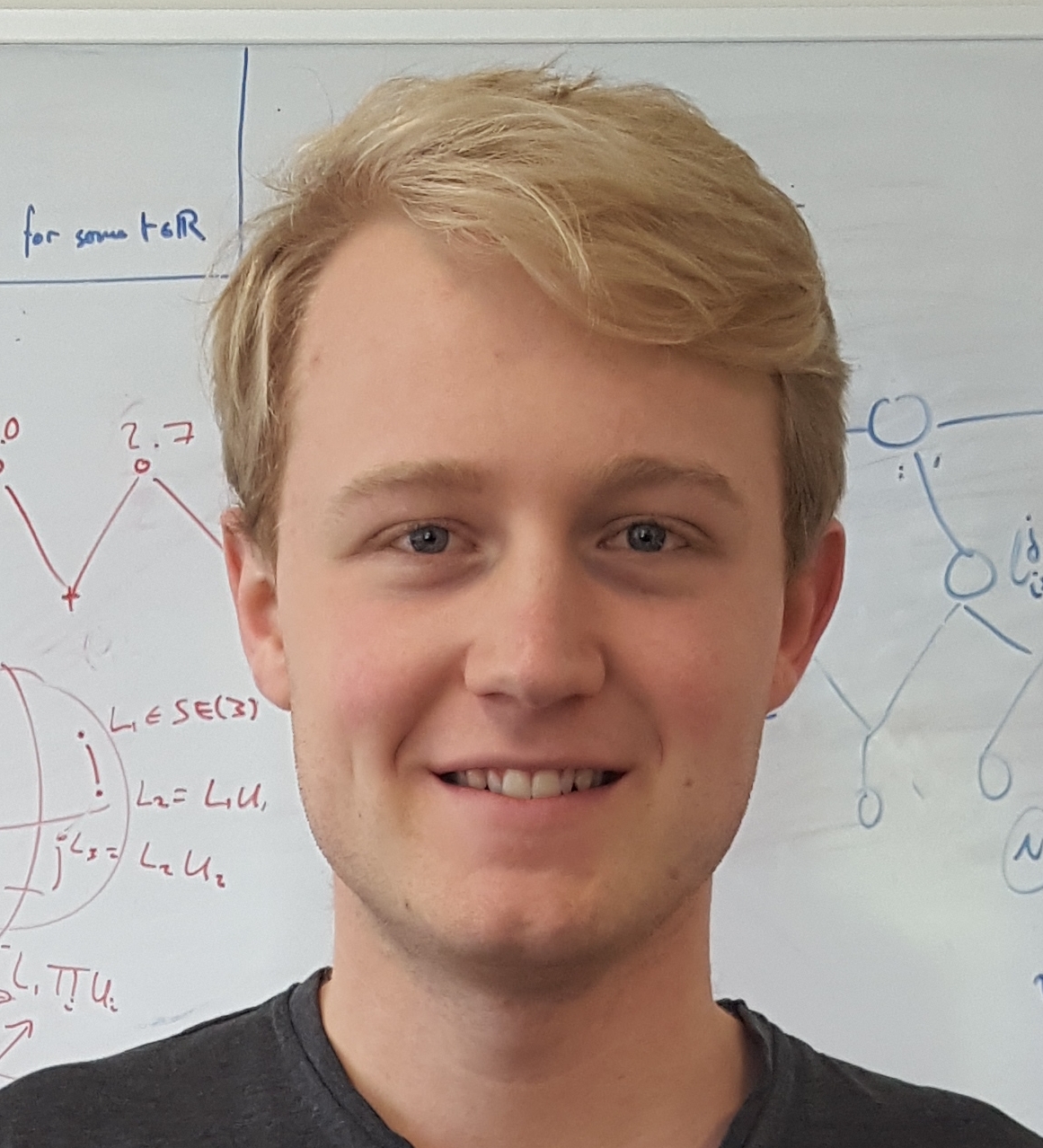}}]{Pieter van Goor}
is a research fellow at the Australian National University.
He received the BEng(R\&D)/BSc (2018) and Ph.D. (Engineering) (2023) degrees from the Australian National University .
His research interests include applications of Lie group symmetries and geometric methods to problems in control and robotics, and in particular visual spatial awareness.
He is an IEEE member.
\end{IEEEbiography}

\begin{IEEEbiography}[{\includegraphics[width=1in,height=1.25in,clip,keepaspectratio]{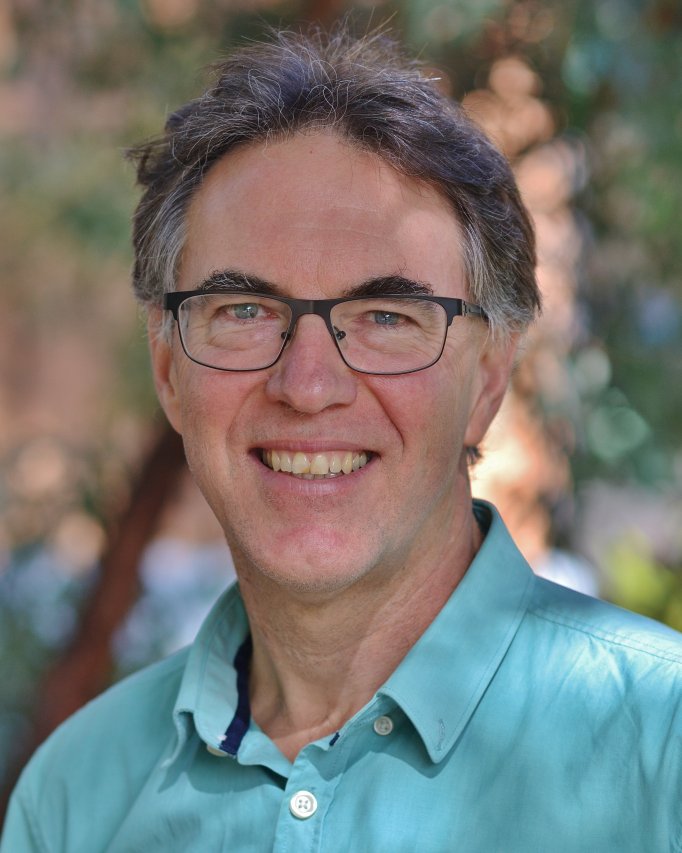}}]{Robert Mahony} (Fellow, IEEE) is a Professor in the School of Engineering at the Australian National University. He is the lead of the Systems Theory and Robotics (STR) group. He received his BSc in 1989 (applied mathematics and geology) and his PhD in 1995 (systems engineering) both from the Australian National University. He is a fellow of the IEEE. His research interests are in non-linear systems theory with applications in robotics and computer vision. He is known for his work in aerial robotics, geometric observer design, robotic vision, and optimisation on matrix manifolds.
\end{IEEEbiography}

\vfill

\end{document}